\newcommand{\trtitle}{Domain-Incremental Continual Learning for Mitigating Bias in Facial Expression and Action Unit Recognition}

\documentclass[journal,twoside,nocompress,letterpaper]{IEEEtran}

\usepackage{cite}

\usepackage{graphicx}
\usepackage{amsfonts}
\usepackage{amsmath}
\usepackage{amssymb}
\usepackage{amsthm}
\usepackage{enumitem}
\usepackage{algorithmic}
\usepackage[table]{xcolor}
\usepackage{array}
\usepackage{placeins}
\usepackage{moresize}
\usepackage[export]{adjustbox}

\ifCLASSOPTIONcompsoc
  \usepackage[caption=false,font=footnotesize,labelfont=sf,textfont=sf]{subfig}
\else
  \usepackage[caption=false,font=footnotesize]{subfig}
\fi
\usepackage{adjustbox}
\usepackage{tcolorbox}
\usepackage{diagbox}

\ifCLASSOPTIONcaptionsoff
  \usepackage[nomarkers]{endfloat}
 \let\MYoriglatexcaption\caption
 \renewcommand{\caption}[2][\relax]{\MYoriglatexcaption[#2]{#2}}
\fi

\usepackage{tabularx}
\usepackage{booktabs}                   
\usepackage{multicol}                   
\usepackage{multirow}                   
\usepackage{makecell}
\usepackage{rotating}                   
\usepackage{enumitem}
\usepackage{url}

\ifCLASSINFOpdf
  \usepackage[pdftex]{thumbpdf}
\else
  \usepackage[dvips]{thumbpdf}
\fi

\newcommand\MYhyperrefoptions{bookmarks=true,bookmarksnumbered=true,
pdfpagemode={UseOutlines},plainpages=false,pdfpagelabels=true,
colorlinks=true,linkcolor={black},citecolor={green},urlcolor={blue},
pdftitle={MitigatingBiasInFERUsingCL},
pdfsubject={Fairness in AI},
pdfauthor={Nikhil Churamani, Ozgur Kara, Hatice Gunes},
pdfkeywords={Continual Learning, Fairness, Bias Mitigation, Facial Expression Recognition, Affective Computing}}
\ifCLASSINFOpdf
\usepackage[\MYhyperrefoptions,pdftex]{hyperref}
\else
\usepackage[\MYhyperrefoptions,breaklinks=true,dvips]{hyperref}
\usepackage{breakurl}
\fi
\usepackage[nolist]{acronym}
\usepackage{balance}
\newcommand{\et}[1]{#1~et~al.}
\newcommand{\nl}{\vspace{1.0mm}{\noindent}}

\begin{document}
\bstctlcite{IEEEexample:BSTcontrol}
{
    \title{\bf\huge\trtitle}
    \author{
    Nikhil~Churamani\IEEEauthorrefmark{1}\href{https://orcid.org/0000-0001-5926-0091}{$^{\includegraphics[scale=0.5]{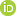}}$},
    Ozgur~Kara\IEEEauthorrefmark{2},
    and~Hatice~Gunes\IEEEauthorrefmark{1}
\IEEEcompsocitemizethanks{
\IEEEcompsocthanksitem 
\IEEEauthorrefmark{1}N.~Churamani and H.~Gunes are with the Department of Computer Science and Technology, University of Cambridge, United Kingdom. \protect\\ \textit{E-mail: \{nikhil.churamani, hatice.gunes\}@cl.cam.ac.uk}

\IEEEcompsocthanksitem 
\IEEEauthorrefmark{2}O.~Kara is with the Electrical \& Electronics Engineering Department at the Bogazici University, Istanbul, Turkey.\protect\\
\textit{E-mail: ozgur.kara@boun.edu.tr}
}

\thanks{\footnotesize 
N.~Churamani is funded by the EPSRC under grant EP/R$513180$/$1$ (ref.~$2107412$). H.~Gunes' work is supported by the EPSRC under grant ref. EP/R$030782$/$1$ and partially by the European Union's Horizon~$2020$ Research and Innovation programme under grant agreement No.~$826232$. O.~Kara contributed to this work while undertaking a summer research study at the Department of Computer Science and Technology, University of Cambridge.\protect\\Authors also thank Prof Lijun Yin from Binghamton University (USA) for providing access to the BP4D Dataset and the relevant race attributes; and Shan Li, Prof Weihong Deng and JunPing Du from the Beijing University of Posts and Telecommunications (China) for providing access to RAF-DB.
}
}}

\IEEEtitleabstractindextext{
    \begin{abstract}
    As \acf{FER} systems become integrated into our daily lives, these systems need to prioritise making \textit{fair} decisions instead of aiming at higher individual accuracy scores. Ranging from surveillance systems to diagnosing mental and emotional health conditions of individuals, these systems need to balance the \textit{accuracy vs fairness} trade-off to make decisions that do not unjustly discriminate against specific under-represented demographic groups. Identifying \textit{bias} as a critical problem in facial analysis systems, different methods have been proposed that aim to mitigate bias both at data and algorithmic levels. In this work, we propose the novel usage of \acf{CL}, in particular, using \acf{Domain-IL} settings, as a potent bias mitigation method to enhance the \textit{fairness} of \ac{FER} systems while guarding against biases arising from skewed data distributions. We compare different non-\ac{CL}-based and \ac{CL}-based methods for their classification \textit{accuracy} and \textit{fairness scores} on expression recognition and \acf{AU} detection tasks using two popular benchmarks, the RAF-DB and BP4D datasets, respectively. Our experimental results show that \ac{CL}-based methods, on average, outperform other popular bias mitigation techniques on both \textit{accuracy} and \textit{fairness} metrics. 
\end{abstract}


}

\maketitle

\IEEEdisplaynontitleabstractindextext
\IEEEpeerreviewmaketitle

\ifCLASSOPTIONcompsoc
\IEEEraisesectionheading{\section{Introduction}\label{sec:introduction}}
\else
\section{Introduction}
\label{sec:introduction}
\fi

\acf{AI} and \ac{ML} systems are increasingly becoming an important part of human life, monitoring and controlling several aspects of our daily lives, with little to no human oversight. From security and surveillance systems that deploy several \ac{ML} models such as face detection and recognition systems~\cite{Feldstein2019AISurveillance}, social media platforms that auto-tag pictures of our friends and family~\cite{Taigman2014DeepFace}, recommender systems that track our digital footprints to show us advertisements of products that we might like to indulge in~\cite{Dey2020Recommendation}, to banking and finance applications that work on credit approvals based on socio-economical backgrounds of individuals, \ac{AI} systems are ubiquitous, making `smart' decisions about several critical aspects of our lives~\cite{Roselli2019,Howard2017Addressing}. It is thus important to ensure that these systems make fair and unbiased decisions to avoid potentially catastrophic consequences that adversely affect individuals~\cite{goodman2017european}. In this work, we focus on one such popular application of \ac{AI} in real life; \acf{FER} systems. 

\ac{FER} systems (see~\cite{Sariyanidi2015Automatic,nott44740,Li12020Deep} for a survey) aim to analyse facial expressions either by encoding facial muscle activity as Facial \acfp{AU}~\cite{ekman1978facial} or determining the emotional state being expressed by an individual~\cite{Ekman1971Constants, Ekman2009}. Analysing large datasets of human faces, annotated for the expressions represented in the images, these models are heavily data-dependent and thus may be prone to \textit{biases} originating from imbalances in the training data distribution. For a large variety of \ac{FER} datasets, attributes such as gender, race, age or skin-colour are implicitly encoded in the data which may also be learnt by a (deep) learning model~\cite{Li2020Deeper}. If these attributes are not balanced across the entire distribution of the dataset, the model may learn to associate such \textit{confounding} attributes with the task of \ac{FER}. For example, if the training data has a disproportionate number of images of Males expressing `Happy' than Females, the model may learn to associate gender with the expression, leading to a lot of `happy female' samples being misclassified.  

While the most effective method for preventing biases in \ac{FER} datasets would be to ensure a balanced and representative data collection, this also turns out to be the most challenging problem. Owing to restrictions with respect to data recording settings, personal preferences, geographic location as well as several social and cultural constraints, it may not always be possible to ensure a balanced data collection. Most recent datasets try to ensure the data collection is fair and unbiased or at the least provide demographic annotations, along with affective labels, that enable researchers to make informed decisions while using these datasets for training \ac{ML} models~\cite{Li2020Deeper}. Yet, to ensure fairness despite the inherent imbalances in data distributions, several methods have been proposed that handle these imbalances at the \textit{pre-processing}, \textit{in-processing} or \textit{post-processing} levels~\cite{yucer2020exploring}. 

Pre-processing methods focus on \textit{strategically sampling} training data, that is, given the distribution of data with respect to a selected demographic attribute, samples belonging to under-represented groups are either over-sampled compared to dominant groups~\cite{elkan2001foundations}, or scaled penalties are applied when a model incorrectly classifies these samples~\cite{Shao2018Deep}. Yet, these methods are not perfect and some bias might still creep in. To handle that, changes to the model architecture or the training regime needs to be made. Algorithm-level or \textit{In-Processing} methods achieve this either by explicitly learning domain-specific information such that this can be discounted from the model's learning later~\cite{Dwork2012Fairness} or they learn to completely discount domain-specific information by omitting these features from the learnt representations~\cite{wang2020towards}. Post-processing methods, on the other hand, are mostly used to quantify bias in trained algorithms~\cite{yucer2020exploring} and offer effective tools to evaluate the \textit{fairness} of an \ac{ML} model.

Interestingly, the underpinning principle behind all the above-mentioned methods is essentially to focus on learning and adapting to the inherent imbalances in data distribution, either by \textit{synthetically} balancing it or adjusting the learning algorithm itself to account for these imbalances. This principle is shared by \acf{CL} methods~\cite{parisi2019continual, LESORT2020CL4R} that also aim to balance learning in the model by being sensitive to shifts in data distributions, ensuring that one particular domain or task does not dominate the model's learning. Their ability to \textit{continually} learn and adapt to novel information, aggregating new knowledge without impacting previously acquired information, may allow them to balance learning across the different learning domains. Domain-Incremental \ac{CL} settings~\cite{van2019three} particularly focus on managing shifts in input data distribution while the task remains the same. This can be considered analogous to solving \ac{FER} tasks where input data belongs to different domains of gender (male, female) and race (black, white, asian). The challenge for \ac{CL} models will thus be to maintain performance on \ac{FER} tasks with respect to one domain while acquiring information about new domains.

Motivated by this notion, we propose the novel use of \ac{CL} as a learning paradigm that is well-suited for developing \textit{fairer} \ac{FER} models that can balance learning with respect to different attributes of gender and race. We formulate expression recognition and \ac{AU} detection across these different domain attributes as a continual learning problem and compare several popular \ac{CL} approaches with state-of-the-art bias mitigation approaches. To the best of our knowledge, this is the first application of \ac{CL} learning as a bias mitigation strategy for facial affect analysis tasks. 
Furthermore, we explore the Domain-Incremental \ac{CL} settings~\cite{van2019three} where the models need to solve the facial analysis tasks across different domains, defined by the demographic attributes of \textit{gender} and \textit{race}. For each attribute, the data is split into different domains, that is, gender annotations are used to define \textit{male} and \textit{female} domains, while race annotations are used to split data into \textit{White/Caucasian, Black/African-American, Asian} and \textit{Latino} domains. We primarily focus on regularisation-based \ac{CL} methods 
as these do not require setting up additional memory or computational resources, allowing a fair and direct comparison with other learning methods that focus on mitigating bias arising due to imbalances in data distributions. Our experimental results show that \ac{CL}-based approaches are, on average, seen to outperform other bias mitigation strategies, both in terms of accuracy as well as fairness scores for both \ac{FER} and \ac{AU} detection tasks across the domain-splits.

%
%

\section{Background and Related Work}
\label{sec:bgrw}

\subsection{Understanding Bias}
Bias, both in human perception and behaviour as well as \ac{ML} algorithms can be characterised as an inclination or prejudice towards a person or a group, that may be considered unfair. This may result from an over or under-exposure of an individual towards a certain group of individuals usually characterised by their gender, racial identity, social or economical background or age, amongst other factors. This exposure results in people considering individuals that share similar characteristics as themselves as ``in-group'' members and others different from them as ``out-group'' members~\cite{tajfel1979integrative}. It is seen that people tend to be \textit{biased} in favour of in-group members, evaluating them more positively on dimensions of judgement while being negatively \textit{biased} or \textit{prejudiced} against out-group members~\cite{Hewstone2002Intergroup}.

Understanding how humans consider in-group and out-group members~\cite{tajfel1979integrative} in their immediate surroundings and base their decisions on aspects such as gender, race or age is important to view \ac{ML} models in the right perspective when applied to real-world settings. Such an understanding will allow researchers to assess what may be considered `fair' and how to achieve such fairness in algorithms.
\subsubsection{\textbf{Bias in Human Perception}}
\label{sec:bias_human}


Following the Perception-Action model of Empathy~\cite{preston2007perception}, an individual's behaviour, particularly their facial expressions and body gestures stimulate a similar neural activation in the observer, enabling them to empathise with and understand their actions, intentions and emotions. \et{Gutsell}~\cite{gutsell2010empathy}, through a series of experiments with multiple participants ($30$ White university students) interacting with in-group (in this case participants with a \textit{Caucasian} ethnic identity) and out-group (excluded from this circle; in this case, \textit{African–Canadian, East-Asian, South-Asian} ethnic identities) members, concluded that such perception-action couplings are reserved only for in-group members. In-group identification, that is, identifying other individuals to be sharing similar characteristics as oneself, causes a \textit{positive association} with them~\cite{Hewstone2002Intergroup}. Furthermore, people have a harder time recognising the faces of out-group members and interpret their facial expressions~ \cite{sporer2001recognizing, elfenbein2002there}.

In the case of \ac{ML} algorithms, we may understand such `inter-group bias' to result from the imbalances in data distributions where certain groups may be considered to constitute `in-group' attributes due to their dominance in the data, while under-represented attributes can be considered as members of the `out-group'. Thus, having witnessed a lot of samples from certain groups, the models are more capable of correctly classifying such samples while performing poorly for the so-called out-group samples.

\subsubsection{\textbf{Bias in Machine Learning}}
\label{sec:bias_ml}
Owing to similar reasons as in the case of human perception, over or under-exposure to experiences (or in this case, data) characterised by specific features, \ac{ML} models are seen to acquire biases that prejudice model performance for one or more data attributes. Facial analysis models particularly are seen to be affected by biases with respect to demographic attributes of gender, race or age, where samples belonging to one group dominate the data distribution. In such situations, the under-represented groups get adversely impacted by the model misclassifying samples from these groups. \et{Buolamwini} in their seminal work~\cite{buolamwini2018gender}, highlighted how popular face recognition algorithms disproportionately misclassified darker females either misgendering them or not being able to detect their faces. Another study by \et{Klare}~\cite{Klare2012Face} highlighted how face recognition algorithms employed by some law-enforcement agencies significantly underperform for people labelled as black or female compared to other demographics. Such biases in critical systems may lead to unnecessary targeting and exploitation of people from under-represented groups, further disadvantaging their opportunities in society.

\subsection{Mitigating Bias in Facial Analyses}
\label{sec:mitigate_bias_face}
The origins of bias in most \ac{ML}-based facial analyses algorithms can be traced back to imbalances in data distributions. Collating balanced datasets that enable a fair evaluation of \ac{ML} models~\cite{Robinson_2020_CVPR_Workshops} despite being the most effective solution towards mitigating such biases, may not be as straightforward to achieve as a varied and diverse subject-pool might not always be available. As a result, several strategies have been proposed for mitigating the effects of bias on the training and evaluation of \ac{ML} algorithms. We use a similar nomenclature as~\cite{yucer2020exploring} to discuss these strategies.

\subsubsection{\textbf{Pre-Processing Approaches}}
The most simplistic of these is achieved by selectively sampling training data in a manner that balances learning. Samples from under-represented domains are over-sampled while dominant domains are under-sampled to balance learning the training data~\cite{Iosifidis2018Dealing,wang2020towards}. This results in the training set to effectively have a balanced data distribution. However, this may not be possible in really small-scale datasets as under-sampling already limited data might not be efficient. An alternative approach is to use data-augmentation techniques to synthetically generate additional data for the under-represented groups~\cite{Han2019Adversarial,Abbasnejad2017Using,CHARTE2015MLSMOTE}, to balance training data distribution.

\subsubsection{\textbf{In-Processing Approaches}}
Another popular approach to mitigate the effects of imbalances in data distributions is to weight model prediction loss differently for the different domain attributes. A weighting factor is applied to the training loss computation based on the occurrence rate for the different classes or domains~\cite{Shao2018Deep, Churamani2020AULACaps, elkan2001foundations} penalising misclassifications for the under-represented groups more than others. This reduces the effect of these imbalances, mitigating biases in learning. 

More recently, several learning strategies have been proposed that, while handing imbalances in data distributions using the above-mentioned techniques, also deal with biases in \ac{ML} models at the algorithm-level. \et{Howard}~\cite{Howard2017Addressing} propose a hierarchical approach that combines outputs from the cloud-based 
Microsoft Emotion API algorithm with a specialised learner, offering a $17.3\%$ improvement in recognition results on a minority class, in this case, children's facial expressions. Other approaches focus on explicitly separating the decision boundaries with respect to different sensitive domain attributes ensuring that imbalances in data with respect to these attributes are not perpetuated while training the model, to achieve \textit{`fairness through awareness'}~\cite{Dwork2012Fairness}. Alternatively, the model can be trained to ignore domain-specific information, making it \textit{unaware} or \textit{blind} towards domain differences and focus only on the task at hand. Adversarial learning has been used to achieve such \textit{`fairness through blindness'}~\cite{wang2020towards} using a min-max training regime that maximises sensitivity towards the task at hand while minimising learning of domain-specific information. \et{Xu}~\cite{xu2020investigating} implement a disentangled approach~\cite{Liu_2018_CVPR} that  uses a similar strategy to mitigate bias with respect to sensitive domain attributes of gender and race for \ac{FER} by ensuring that the feature representations learnt by the model do not contain any domain-specific information. The model is split into two parts with a shared feature extraction sub-network. The first part focuses on facial expression analysis, while the other part consists of separate branches for each domain, designed to suppress domain-specific information. 

\subsubsection{\textbf{Post-Processing Approaches}}
Despite several methods proposed for training \textit{fair} \ac{ML} systems, as described above, it may not always be possible to completely eradicate bias in the model. In such cases, it is still important to examine whether a model is biased and quantify the bias to mitigate it and make fairer decisions. Post-processing approaches (see~\cite{yucer2020exploring,Srinivas2019Face} for a general discussion) focus on quantifying bias in existing algorithms and attempt to counter the effects on classification tasks.

\begin{figure*}
    \centering
    %
    \includegraphics[width=0.65\textwidth]{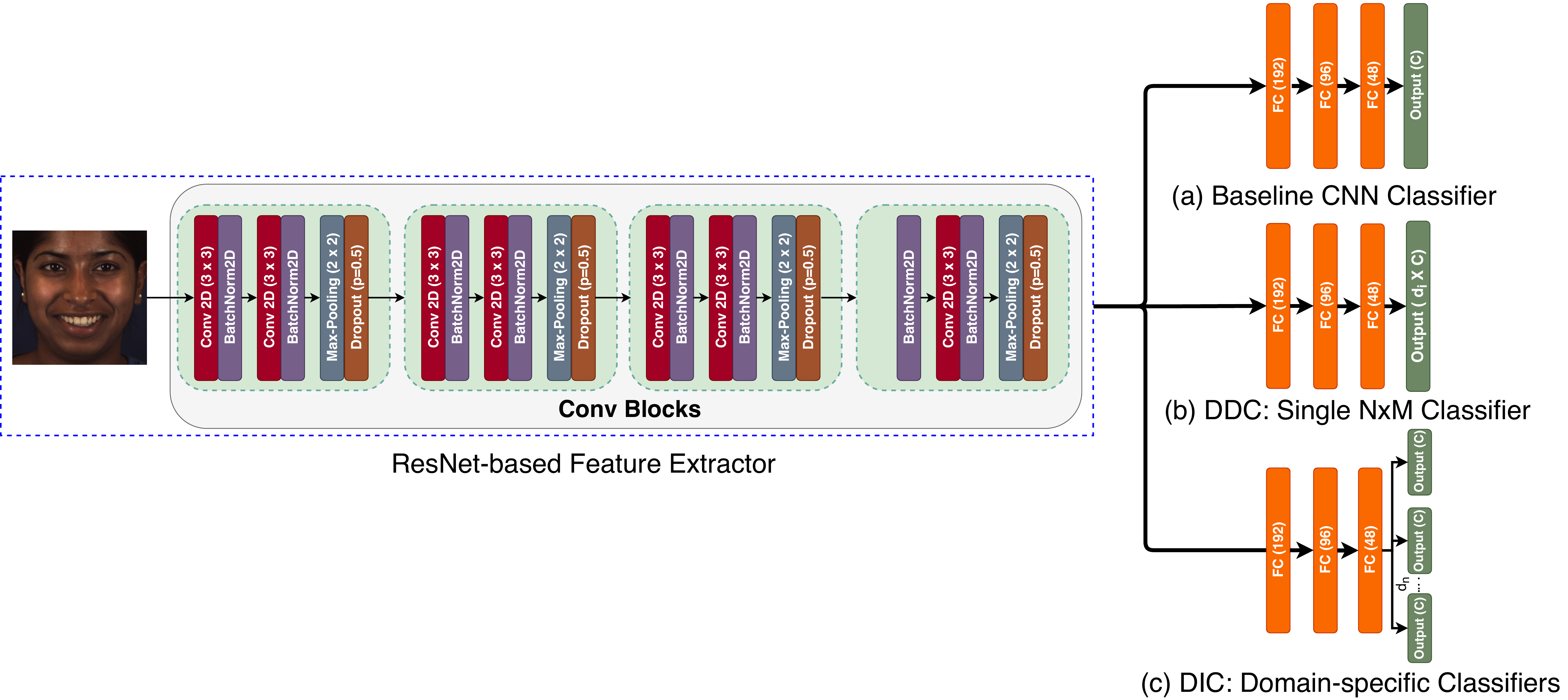}
    \vspace{-2mm}\caption{Model Architectures for (a)~the Baseline CNN,~(b)~\acf{DDC}~\cite{wang2020towards},~(c)~\acf{DIC}~\cite{wang2020towards}.
    The Baseline CNN is also used to implement all \ac{CL} methods for a fair comparison.}
    \label{fig:model_architectures}
\end{figure*}

\subsection{Continual Learning}
\label{sec:cl_rw}
Learning to detect and manage shifts in data distributions, \acf{CL} methods (see~\cite{parisi2019continual,LESORT2020CL4R} for an overview) can effectively learn with incrementally acquired data, offering an improvement over traditional \ac{ML} models, especially for real-world application. 

Typically, \ac{CL} models are evaluated on $3$ different learning scenarios~\cite{van2019three}. The first scenario is termed as \textit{\acf{Task-IL}} where the model incrementally learns to solve several tasks, explicitly being informed about the task identity. Learning is split into different tasks, each corresponding to learning some sub-tasks or classes. The model is evaluated on its ability to preserve its knowledge across several tasks. The second scenario focuses on \textit{\acf{Domain-IL}} where the task to be learnt by the model does not change but the input data distribution changes. While the model still needs to solve the same task but the inherent data distribution shifts and the model is evaluated on its ability to manage such a shift. The third and the most complex learning scenario is the \textit{\acf{Class-IL}} scenario where the model needs to learn a new class without being given any information on the tasks. The model incrementally learns one class at a time, sequentially receiving input data for only that class. 

In recent years, several \ac{CL} approaches have been proposed that employ (deep) \ac{ML} architectures and equip them with learning capabilities such that they can incrementally integrate novel information while preserving past knowledge~\cite{parisi2019continual}. The most common and straightforward approach to achieve this is by regulating model updates in a manner that enables the preservation of knowledge. Such \textit{regularisation-based} methods minimise \textit{destructive interference} by freezing those parts of the model that are most sensitive to previous tasks~\cite{fernando2017pathnet} and updating the rest of the model, selectively. Alternatively, weight-update constraints and penalties are applied that discourage changes in network parameters that deteriorate model performance on previous tasks~\cite{li2017learning, kirkpatrick2017overcoming}. A priority or importance term may also be applied to network parameters based on their relevance to a given task and only those parameters are allowed to be updated which have lower importance~\cite{zenke2017continual}. 
Despite the competitive performance of regularisation-based methods, they become computationally expensive as the number of tasks or classes grow, limiting their performance and applicability in \ac{Task-IL} (in extreme cases) and \ac{Class-IL} scenarios.

Other \ac{CL}-based approaches include rehearsal-based methods~\cite{Robins1993Rehearsal} that aim to simulate offline batch-learning based settings by either physically storing previously encountered data samples; commonly known as \textit{\acf{NR}}~\cite{Hsu18_EvalCL}, or learning a generative or probabilistic model that learns data statistics to simulate \textit{pseudo-samples} for previously seen tasks~\cite{Robins1995, shin2017continual,churamani2020clifer}. Yet, as the number of tasks increase, it becomes extremely difficult to train these models. Furthermore, additional memory and computational resources need to be allocated to either store data samples or generate simulated pseudo-samples making it challenging to implement these approaches. 

In this work, we focus on regularisation-based methods evaluated under \ac{Domain-IL} settings where these models are required to learn to solve expression recognition and \ac{AU} detection tasks across different domains of gender and race. 

\section{Methodology}
\label{sec:method}
In order to understand bias in \ac{FER} algorithms, it is important to determine how the implicit data distribution affects model performance. For this, we need to understand which domain attributes dominate the data and how an algorithm performs with respect to these attributes. In this section, we present the problem formulation, the learning scenario as well as the different methods employed in this work, comparing them with popular \ac{CL}-based methods.

\subsection{Problem Formulation}
We aim to measure the variance in model performance on a specific task with respect to \textit{gender} and \textit{race} as the different domain attributes and compare model performances for expression recognition and \ac{AU} detection. Given a set of input images $x_i$ with task labels $y_i$ and domain label $d_i$, we wish to determine how the performance of an algorithm $\mathcal{A}(x_i|y_i, d_i)$, varies with respect to different variations of the domain label $d_i$. 



To enable a fair comparison between the different bias mitigation methods, for our experiments, we implement the same ResNet architecture-based~\cite{He2016resnet} \acs{CNN} model for all the methods consisting of $4$ convolutional blocks, each with $2$ conv layers, a max-pooling layer and implementing drop-out with batch-normalisation. The output of the last conv block is connected to three dense layers and a classification layer making model predictions. ReLU activation is used for each conv as well as dense layer. The same architecture is used to implement all the approaches compared in this work (see Fig.~\ref{fig:model_architectures}) with the exception of the Disentangled Approach for which the results from the original paper~\cite{xu2020investigating} are used directly for comparison purposes.

\subsubsection{\textbf{The Baseline}}
For baseline evaluations, we split the dataset into different subsets based on the domain attributes. For example, for gender, the datasets are split into male and female splits and model performance is reported when trained incrementally on these data splits. This is sometimes also referred to as \textit{finetuning}~\cite{aljundi2018memory}. The model (see Fig.~\ref{fig:model_architectures}a), without any explicit mechanism to preserve knowledge, is expected to suffer from forgetting old tasks while preference is given to the new tasks. 

\subsubsection{\textbf{Off-line Training}}
Providing another baseline evaluation, the above-described  \ac{CNN} model (see Fig.~\ref{fig:model_architectures}a) is trained on all the training data, \textit{off-line}, at once but its performance scores are reported individually on domain-specific test-splits. Off-line training provides a fair comparison with traditional \ac{ML}-based learning models and is a popularly used benchmark for evaluating the performance of \ac{CL}-based methods.

\subsection{Non-\acs{CL}-based Bias Mitigation Strategies}
\label{subsec:traditionalbias}
Here, we investigate some of the popular bias mitigation strategies that are found in the literature and implement $4$ different methods for comparison. We group them under `non-\ac{CL}-based' strategies to differentiate them from our baselines as well as the \ac{CL} approaches.

\subsubsection{\textbf{\acf{DDC}}}
A popular method for mitigating bias is to focus on achieving `fairness through awareness'~\cite{Dwork2012Fairness} where information about sensitive attributes (or domains) is explicitly learnt in feature encodings. This information later allows models to account for bias in learning by being more `aware'. One way to achieve this to create an $N\times M$-way discriminative classifier where $N$ denotes the number of domains and $M$ is the number of classes to be learnt~\cite{wang2020towards}. For example, for \ac{FER} classifying $7$ different expression classes for samples encoding $3$ different race labels, a classifier is used with each output unit corresponding to a unique expression-race label pair (in this case, $7\times3=21$ label pairs). This allows the model to be more `aware' of the different domains in order to learn discriminative features for each of them. For our experiments, we use the same model architecture (see Fig~\ref{fig:model_architectures}b) only replacing the output layer. 

\subsubsection{\textbf{\acf{DIC}}}
A major concern with the \ac{DDC} method is that the network may implicitly learn decision boundaries within the same class across different domains. This may be redundant as, despite the different domain attributions, the class-boundaries may remain the same and the network may be unnecessarily penalised due to incorrect domain predictions even if it predicts the task correctly. \et{Wang}~\cite{wang2020towards} offer a solution to this by training separate classifiers for each domain, sharing the feature extraction layers. For our experiments, we make use of the same model architecture (see Fig.~\ref{fig:model_architectures}c), connecting separate dense-layered classifiers for each of the domains. The \ac{DIC} model consists of different \textit{heads}, each consisting of the same number of output units but corresponding to different domain attribute labels.

\subsubsection{\textbf{\acf{SS}}}

A simple approach for handling bias arising from imbalanced data distributions is to strategically sample data~\cite{elkan2001foundations} for each domain-class mapping such that the resultant data distribution `appears' to be balanced. Samples from under-represented distributions can be sampled more often during training or equivalently, prediction loss can be appropriately weighted to account for the under-represented classes. For our experiments, samples ($s$) for each of the $N$ domains ($d_i$) are assigned a weight $w_i$ \textit{inversely proportional} to the rate of occurrence of samples for that domain, scaling the loss function to appropriately account for imbalances in the training set distribution. The scaled cross-entropy loss function is given as:

{\small
\begin{equation}
L(\mathbf{y}, \mathbf{\hat{y}}) = -\frac{1}{N}\sum_{i=1}^{N} w_i \sum_{s=1}^{S}y_s^{(i)}\log{\hat{y}_s^{(i)}}
\end{equation}
}\subsubsection{\textbf{\acf{DA}}}

\et{Xu}~\cite{xu2020investigating} implement the \acf{DA} approach of~\cite{Liu_2018_CVPR} for facial expression recognition. This approach ensures that the feature representations learnt by the model do not contain any domain-specific information. The two sub-parts of the model focus on analysing facial expressions while learning to suppress domain-specific information. 
For our experiments, we sue the results from the original paper~\cite{xu2020investigating} for comparison.

\subsection{Continual Learning Approaches}
\label{subsec:biascl}

Domain-Incremental \ac{CL} deals with scenarios where the structure of the tasks remains the same albeit with the input distribution is changing~\cite{van2019three}. For our experiments, we model the tasks of expression recognition and \ac{AU} detection in a domain-incremental manner where the models learn to solve these tasks as the input data distributions change with respect to domain attributes of gender and race.  For example, in the case of gender, the models first learn to classify expression classes or predict activated \acp{AU} for `male' samples and then, sequentially, learn to solve these tasks for `female' samples (or vice-versa), without forgetting the previous task.
In our experiments, each approach is implemented using the Baseline \ac{CNN} architecture as shown in Fig~\ref{fig:model_architectures}a. All implementations were based on the \ac{CL} code-benchmarks provided by~\cite{Hsu18_EvalCL,van2019three}.

\subsubsection{\textbf{\ac{EWC}}}
The \ac{EWC} approach, as proposed by \et{Kirkpatrick}~\cite{kirkpatrick2017overcoming} imposes a quadratic penalty on parameter updates between old and new tasks in order to avoid forgetting previously learnt information. For each parameter $\theta$, its relevance is calculated with respect to a task's training data $\mathcal{D}$, modelled as the posterior distribution $p(\theta | \mathcal{D})$. Thus, for two data distributions $\mathcal{D}_A$ and $\mathcal{D}_B$, corresponding to two independent tasks $A$ and $B$, according to Bayes' rule, the posterior probability is given as:

{\small
\begin{equation}
\log p(\theta | \mathcal{D}) = \log p(\mathcal{D}_B | \theta) + \log p(\theta | \mathcal{D}_A) - \log p(\mathcal{D}_B),
\end{equation}
}such that $\log p(\theta | \mathcal{D}_A)$ embeds all information about previously learnt tasks. As this term becomes intractable, Laplace approximation is used to approximate it as a Gaussian Distribution with its mean given by parameters $\theta_{A}^*$ (referring to parameters of task $A$) and the importance of the parameters determined by the diagonal of the Fischer Information Matrix. The loss function for the \ac{EWC} method thus becomes:

{\small
\begin{equation}
L(\theta) = L_B (\theta) + \frac{1}{2} \lambda \sum_i F_i (\theta_i - \theta_{A,i}^*)^2,
\end{equation}
}where $L_B$ is the loss for task $B$,  $\lambda$ is the regularisation coefficient that determines the relevance of old tasks with respect to the new one, $i$ denotes the index of the parameter $\theta$ and $F_i$ is the $i^{th}$ diagonal element of the Fischer Matrix. 

\subsubsection{\textbf{\ac{EWC}-Online}} 
A disadvantage for the \ac{EWC} method is that as the number of tasks increase, the number of quadratic terms in the regularisation term grows linearly.  To handle this, \et{Schwarz}~\cite{schwarz2018progress} proposed a modification to \ac{EWC} where instead of many quadratic terms, a single quadratic penalty is applied, determined by a running sum of the Fischer Information Matrices of the previous tasks. Thus, the updated regularisation term of the proposed \ac{EWC}-online approach is given as:

{\small
\begin{equation}
L_{reg}^T = \sum_i \tilde{F}_{i}^{(T-1)} (\theta_i - \theta_{i}^{(T-1)})^2,
\end{equation}
}where $\theta_{i}^{(T-1)}$ is the $i^{th}$ parameter after learning task $T - 1$ and $ \tilde{F}_{i}^{(T-1)}$ is the running sum of the diagonal elements of the Fischer Matrices of all previous tasks calculated as:

{\small
\begin{equation}
 \tilde{F}_{i}^{(T)} = \gamma  \tilde{F}_{i}^{(T-1)} + F_i^{T},
\end{equation}
}where $\gamma$ controls the contribution of previously learnt tasks.

\begin{figure}[t]
    \centering
    \hspace*{-2mm}\subfloat[Gender\label{fig:rafdb_gender}]{\includegraphics[width=0.25\textwidth]{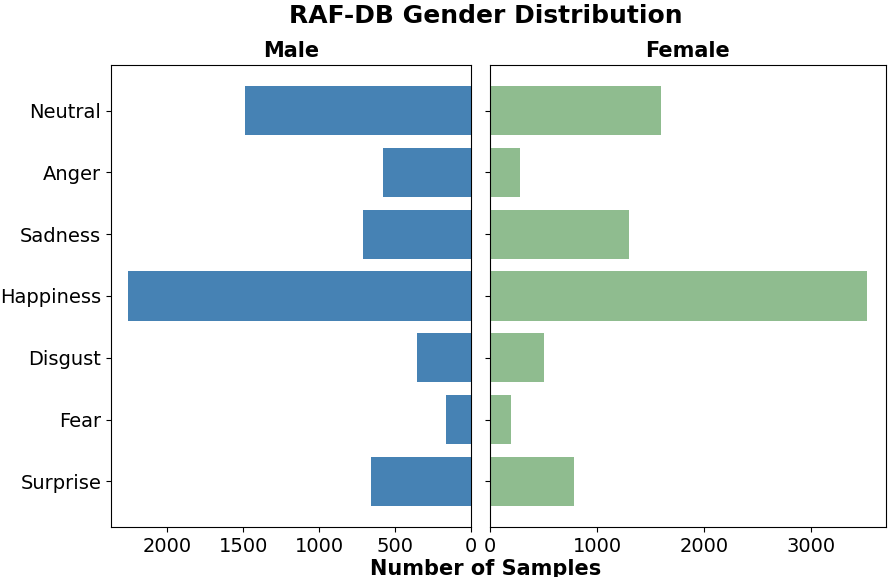}}
    \subfloat[Race\label{fig:rafdb_race}]{\includegraphics[width=0.25\textwidth]{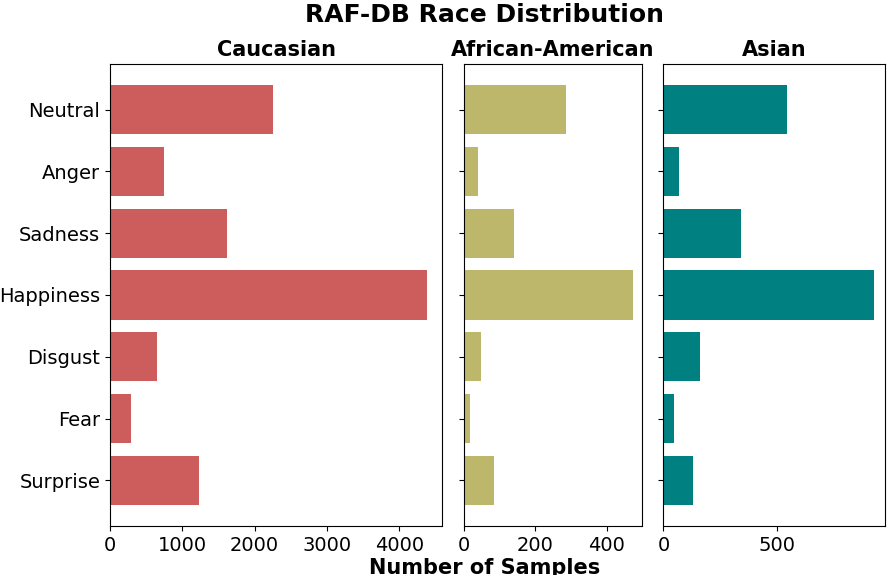}}
    \caption{RAF-DB data distribution for Gender and Race Attributes.}
    \label{fig:rafdb_distribution}
\end{figure}

\subsubsection{\textbf{\acf{SI}}}
Similar to \ac{EWC}, this approach also penalises changes to relevant weight parameters (synapses) in a manner that new tasks can be learnt without forgetting the old~\cite{zenke2017continual}. To alleviate catastrophic forgetting, the importance for solving a learned task is computed for each individual synapses and changes in the most important synapses are discouraged. A modified cost function $L_n^*$ is used with a surrogate loss term which approximates the summed loss functions of all the previous tasks $L_o^*$:

{\small
\begin{equation}
L_n^* = L_n + c\sum_i \Omega_k^n (\theta_k^* - \theta_k)^2,
\end{equation}
}where $\theta_k$ represents the parameters for the new task,  $\theta_k^*$ represents the parameters at the end of the previous task, $\Omega_k^n$ is the parameter regulation strength and $c$ is the weighting factor balancing new vs. old learning.

\begin{table}[t]
    \centering
	\setlength\tabcolsep{3.0pt}

    {
    \scriptsize
    \caption{Regularisation Coefficient values for FER experiments with the RAF-DB dataset~$\mathit{(\times 10^3})$.}    
    \label{tab:lambda_rafdb}
    \begin{tabular}{c|cc|cc}
        \toprule
        
         \multirow{2}{*}{\textbf{Method}} & \multicolumn{2}{c|}{\textbf{W/O Data-augmentation}} & \multicolumn{2}{c}{\textbf{W/ Data-augmentation}}\\
        \cmidrule{2-5}
  													& 		\textit{\textbf{Gender}} 		& 		\textit{\textbf{Race}}		&		 
  															\textit{\textbf{Gender}}		&		\textit{\textbf{Race}} 	 \\ \midrule 
    
        \acs{EWC} ($\lambda$) & 5  & 10 
        & 10 & 5 
        \\
        \acs{EWC}-Online ($\gamma$) & 10 & 1 
        & 10 & 5 
        \\
        \acs{SI} ($c$)& 1  & 10 
        & 1 & 5 
        \\
        \acs{MAS} ($\lambda$) & 0.5 & 5 
        & 1& 0.2 
        \\
		\acs{NR} ($\lambda$) & 0.4  & 1.1 
		& 0.1 & 1.1 
		\\   \bottomrule
    \end{tabular}}
\end{table}

\subsubsection{\textbf{\acf{MAS}}}
Similar to \ac{EWC} and \ac{SI}, \ac{MAS} also calculates the importance of each parameter although by looking at the sensitivity of the output function instead of the loss~\cite{aljundi2018memory}. For each new sample, \ac{MAS} updates the importance of each parameter by evaluating how sensitive the model prediction is to the changes in that parameter. Parameters that have the most impact on model predictions are given high importance and changes to these parameters are penalised. Different from \ac{EWC} and \ac{SI}, parameter importance is computed only using unlabelled data by measuring changes in model performance. For each new task ($T_n$), in addition to the task-loss ($L_n(\theta)$), changes to parameters important for previous tasks are penalised:

{\small
\begin{equation}
L(\theta) = L_n(\theta) + \lambda\sum_{i,j} \Omega_{i,j} (\theta_{i,j} - \theta_{i,j}^*)^2,
\end{equation}
}where $\lambda$ is the hyperparameter balancing new vs. old task losses, and $\theta^*$ denotes the old network parameters.

\subsubsection{\textbf{\acf{NR}}}
For the \acf{NR} approach,  we implement a straightforward rehearsal-based method that combines new data with previously seen data while training the model.~\cite{Hsu18_EvalCL}. A small replay buffer is implemented to randomly store a fraction of previously seen data samples that can be replayed to the model. Each mini-batch of data is constructed using an equal number of samples from the new as well as previously seen data. This interleaving of data pertaining to previously learnt tasks with new data ensures that old knowledge is not overwritten by new data.


\begin{figure}[t]
	\centering
	\subfloat[Gender\label{fig:bp4d_gender}]{\includegraphics[width=0.245\textwidth]{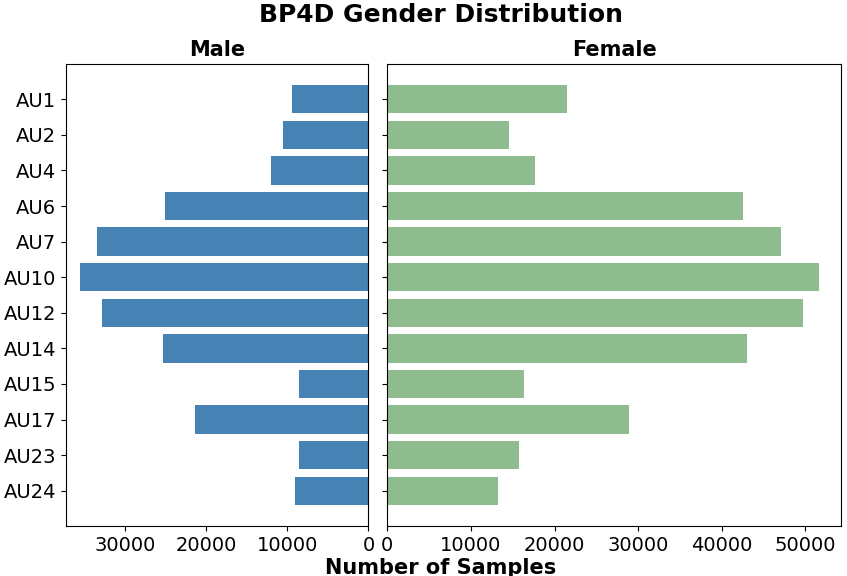}}
	\subfloat[Race\label{fig:bp4d_race}]{\includegraphics[width=0.25\textwidth]{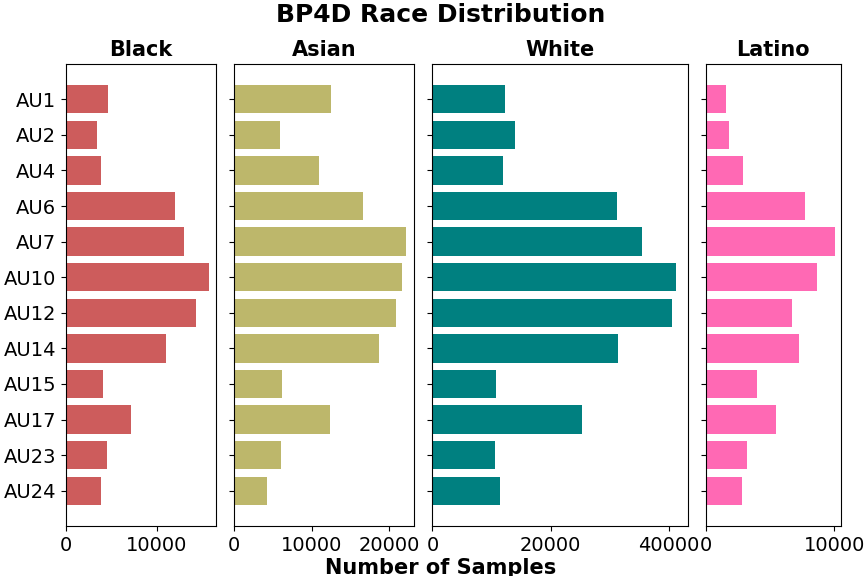}}
	\caption{BP4D data distribution for Gender and Race Attributes.}
    \label{fig:bp4d_distribution}
\end{figure}

\begin{table}[t]
    \centering
	\setlength\tabcolsep{3.0pt}

    {
    \scriptsize
    \caption{Regularisation Coefficient values for \ac{AU} detection experiments with the BP4D dataset~$\mathit{(\times 10^3})$.}    
    \label{tab:lambda_bp4d}
    \begin{tabular}{c|cc|cc}
        \toprule
         \multirow{2}{*}{\textbf{Method}} & \multicolumn{2}{c|}{\textbf{W/O Data-augmentation}} & \multicolumn{2}{c}{\textbf{W/ Data-augmentation}}\\
        \cmidrule{2-5}
        															& 		\textit{\textbf{Gender}} 		& 		\textit{\textbf{Race}}		&			\textit{\textbf{Gender}}		&		\textit{\textbf{Race}} 	 \\ \midrule
        															\
        \acs{EWC} ($\lambda$)					& 			5  			& 			5			& 				5 				& 			1  		 \\
        \acs{EWC}-Online ($\gamma$)	& 		0.05 			& 		0.5 			& 			0.05			& 		0.01     \\
        \acs{SI} ($c$)									&			5  			& 			5 			&			0.05 			& 		0.01 	 \\
        \acs{MAS} ($\lambda$)					& 			1 				& 		0.5 			& 			0.1 				& 			1		 \\
        \acs{NR} ($\lambda$)						& 		0.4 	 			& 		0.1 			& 			0.1  				&		 0.1  	 \\ \bottomrule
    \end{tabular}}
\end{table}

\section{Experiment Set-Up}
\label{sec:setup}
\subsection{Datasets}
For evaluating the different bias mitigation strategies and comparing them with \ac{CL}-based methods, we use two popular benchmark datasets; the RAF-DB dataset for \ac{FER} \textit{in-the-wild} and the BP4D dataset recorded for \ac{AU} detection in controlled settings . These datasets are selected due to (i) the diversity in their data acquisition settings, (ii) providing labels not only for  expression/AU recognition but also gender and race attributes, and (iii) containing notable imbalances in the data distributions with respect to class and domain attribute labels. These factors make the RAF-DB and the BP4D datasets a good choice for our evaluation.

\subsubsection{\textbf{RAF-DB Dataset}}
The \acs{RAF-DB} dataset~\cite{li2017reliable} consists of $\approx15K$ facial images labelled for six expression classes namely, \textit{Surprise, Fear, Disgust, Happy, Sad} and \textit{Anger} along with \textit{Neutral} to denote absence of any expression. Additionally, it provides demographic attribute labels such as gender (Male, Female, Unsure) and race (Caucasian, African-American, Asian) labels. 
For our experiments, we split the dataset using multiple grouping strategies based on the gender and race Labels. 
For gender-based grouping, we exclude images labelled as `Unsure' and only use the `Male' and `Female' samples.  
As shown in Fig.~\ref{fig:rafdb_distribution}, not only is the dataset imbalanced with respect to the different expression categories,  there exist stark imbalances with respect to different demographic attributes as well. The majority of the samples in the training set represent the ``Happy'' expression class and belong to ``Female'' and ``Caucasian'' categories.  


\subsubsection{\textbf{BP4D Dataset}}
The \acs{BP4D} dataset~\cite{ZHANG2014BP4D} consists of video sequences from $41$ subjects performing $8$ different affective tasks to elicit emotional reactions. Each video is annotated frame-wise for the occurrence and intensity of the activated \acsp{AU}. In our experiments, we only use occurrence labels for $12$ most frequent \ac{AU} resulting in $\approx 150K$ labelled frames, in total. Other than the frame-wise \acs{AU} labels, demographic attribute labels for gender (Male, Female) and race (Black, White, Latino, Asian) have been provided to us, specifically for this research. Fig.~\ref{fig:bp4d_distribution} shows the data distribution of the \acs{BP4D} dataset for the $12$ \acs{AU} labels with respect to the gender and race attributes. As can be seen, the majority of the samples in the dataset represent ``White'' and ``Female'' attribute labels.

\subsection{Pre-processing and Data-Augmentation}
Both RAF-DB and BP4D datasets provide face-centred RGB images which are resized to $(100 \times 100 \times 3)$ and normalised to be used as input for all the models. Training deep neural networks requires a lot of training data for each of the classes to be learnt. Due to the inherent imbalances in the dataset with respect to the different expression classes (RAF-DB) or the \ac{AU} labels (BP4D), we increase the overall training data by performing data-augmentation by randomly ($p=0.5$) flipping images horizontally to create additional samples.  For each experiment, we present the results \textit{with} and \textit{without} data-augmentation separately, for clarity.

\subsection{Experiment Settings}
\subsubsection{\textbf{Evaluation Metrics}}
To compare the different methods on their ability to balance classification performance within individual domain-splits while remaining consistent across the domains, we evaluate them both in terms of their accuracy scores as well as \textit{fairness}. Furthermore, for the \ac{CL} methods, we also report \acf{CF} scores, measuring the ability of the models to maintain performance on previously seen tasks while learning new tasks. 
 
\nl\textbf{Accuracy (Acc)}: Accuracy is defined as the fraction of correctly classified samples. Given that TP = True Positives, FP = False Positives, TN = True Negatives and FN = False Negatives, Accuracy (Acc) can be computed as:

{\small
\begin{equation}
Acc = \frac{TP + TN}{TP + FP + TN + FN}
\end{equation}
}In our experiments, we report accuracy scores separately for different gender and race attributes to highlight differences in model performance for these domains, underlining bias in the models' performance.

\begin{table*}[t]
	\centering
	{
	\scriptsize
	\caption{\textbf{Experiment 1:} Gender-wise Accuracy and Fairness Scores on RAF-DB dataset. Accuracy scores are reported after training the models on both Male and Female subsets. \textbf{Bold} values denote best while [\textit{bracketed}] denote second-best values for each column.} 
	\label{tab:gender_raf_db_fairness}
	\begin{tabular}{c|c|c|c|c|c|c}\toprule
	
		\multirow{2}{*}{\textbf{Method}} & \multicolumn{2}{c|}{\textbf{Accuracy W/O Data-Augmentation}} & \multicolumn{1}{c|}{\multirow{2}{*}{\textbf{Fairness}}} & \multicolumn{2}{c|}{\textbf{Accuracy W/ Data-Augmentation}} & \multicolumn{1}{c}{\multirow{2}{*}{\textbf{Fairness}}} \\ \cmidrule{2-3} \cmidrule{5-6}
	
		 & \multicolumn{1}{c|}{\textit{\textbf{Male}}} & \multicolumn{1}{c|}{\textit{\textbf{Female}}} & \multicolumn{1}{c|}{} & \multicolumn{1}{c|}{\textit{\textbf{Male}}} & \multicolumn{1}{c|}{\textit{\textbf{Female}}} & \multicolumn{1}{c}{} \\ \midrule
		 
		Baseline & 0.596$\pm$0.025 & 0.714$\pm$0.014 &  0.834 & 0.596$\pm$0.017 & 0.730$\pm$0.008 &  0.816 \\ 
		Offline  & 0.704$\pm$0.011 & 0.746$\pm$0.007 & 0.944 & 0.724$\pm$0.006 & 0.759$\pm$0.007 &  0.954\\ \midrule

		\multicolumn{7}{c}{\textbf{Non-\ac{CL}-based Bias Mitigation Methods}}\\ \midrule
		
		\ac{DDC}~\cite{Dwork2012Fairness} & 0.699$\pm$0.013 & 0.722$\pm$ 0.008 & 0.968 & 0.717$\pm$0.013 & 0.746$\pm$0.007 &  0.961 \\ 
		
		\ac{DIC}~\cite{wang2020towards} & 0.698$\pm$0.014 & 0.744$\pm$ 0.006 &  0.938 & 0.729$\pm$0.008 & 0.758$\pm$0.002 &  0.962  \\ 
		\ac{SS}~\cite{elkan2001foundations} & 0.716$\pm$0.010 & [\textit{0.750$\pm$0.008}] & 0.955 & 0.729$\pm$0.013 & [\textit{0.764$\pm$0.011}] &  0.954 \\ 
			
		\ac{DA}~\cite{xu2020investigating} & 0.625 & 0.610  & 0.975 & [\textit{0.742}] & 0.744 &  [\textit{0.997}] \\ \midrule
	
		\multicolumn{7}{c}{\textbf{Continual Learning Methods}}\\ \midrule
		
		\ac{EWC}~\cite{kirkpatrick2017overcoming} & \cellcolor{gray!25}\textbf{0.723$\pm$ 0.006} & 0.744$\pm$0.006 & 0.972 & 0.735$\pm$0.007 & 0.748$\pm$0.012 & 0.983 \\ 

		\ac{EWC}-Online~\cite{schwarz2018progress} & 0.721$\pm$ 0.008 & 0.743$\pm$0.006 & 0.970 & 0.736$\pm$0.003 & 0.756$\pm$0.010 & 0.974 \\ 
		
		\ac{SI}~\cite{zenke2017continual} &  0.718$\pm$ 0.007 & 0.725$\pm$0.004 & \cellcolor{gray!25}\textbf{0.990} & 0.739$\pm$0.008 & 0.739$\pm$0.005 & \cellcolor{gray!25}\textbf{0.999} \\ 
	
		\ac{MAS}~\cite{aljundi2018memory} &0.721$\pm$ 0.008 & 0.735$\pm$0.012 & [\textit{0.980}] & \cellcolor{gray!25}\textbf{0.745$\pm$0.006} & 0.753$\pm$0.009 & 0.990 \\ 
		
		\ac{NR}~\cite{hsu2018re} & [\textit{0.722$\pm$ 0.001}] & \cellcolor{gray!25}\textbf{0.778$\pm$0.006} & 0.928 & 0.738$\pm$0.004 & \cellcolor{gray!25}\textbf{0.799$\pm$0.005} & 0.923 \\ \bottomrule

\end{tabular}}
\end{table*}

\begin{table*}[t]
	\centering
	\caption{\textbf{Experiment 1:} \acf{CF} and Overall Accuracy (previous tasks) after each task for Gender-ordered learning on RAF-DB dataset. \textbf{Bold} values denote the best while [\textit{bracketed}] denote second-best values for each column.}
	\label{tab:gender_raf_db_cf}
	{
	\scriptsize
	\begin{tabular}{c|c|c|c|c|c|c|c|c}\toprule

		\multirow{3}{*}{\textbf{Method}} & \multicolumn{4}{c|}{\textbf{W/O Data-Augmentation}} & \multicolumn{4}{c}{\textbf{W/ Data-Augmentation}} \\ \cmidrule{2-9} 
		
		& \multicolumn{2}{c|}{Male} & \multicolumn{2}{c|}{Female} & \multicolumn{2}{c|}{Male} & \multicolumn{2}{c}{Female} \\ \cmidrule{2-9} 
 
		& Acc. & CF & Acc. & CF & Acc. & CF & Acc. & CF \\ \midrule
		
		\ac{EWC}~\cite{kirkpatrick2017overcoming} & \cellcolor{gray!25}\textbf{0.730} & X & [\textit{0.734}] & \cellcolor{gray!25}\textbf{-0.028} & \cellcolor{gray!25}\textbf{0.746} & X & 0.742 & \cellcolor{gray!25}\textbf{-0.032} \\ 
		
		\ac{EWC} Online~\cite{schwarz2018progress} & 0.728 & X & 0.733 & -0.015 & [\textit{0.745}] & X & 0.746 & [\textit{-0.024}] \\ 
		
		\ac{SI}~\cite{zenke2017continual} & 0.721 & X & 0.728 & -0.003 & 0.741 & X & 0.738 & -0.004 \\ 
		
		\ac{MAS}~\cite{aljundi2018memory} & 0.721 & X & 0.731 &[\textit{-0.024}] & 0.743 & X & [\textit{0.749}] & -0.013 \\ 
		
		\ac{NR}~\cite{hsu2018re} & [\textit{0.729}] & X & \cellcolor{gray!25}\textbf{0.753} & -0.010 & 0.736 & X & \cellcolor{gray!25}\textbf{0.772} & -0.010 \\ \bottomrule
		
\end{tabular}}
\end{table*}

\nl\textbf{Fairness Measure ($\mathcal{F}$)}: To evaluate different approaches for their \textit{fairness} with respect to model performance for gender and race attributes, we use the `equal opportunity' definition of \textit{fairness}, as proposed by \et{Hardt}~\cite{hardt2016equality}.  

Let \textbf{x}, \textbf{y}, \textbf{ŷ} be the variables denoting input, ground truth label and the predicted label, respectively, $s\in S_i$ be the sensitive (domain) attribute (for example, $(S_i = \{$male, female$\})$,  $f$ be a function computing the accuracy score for a given sensitive attribute $s$ and $d$ be the dominant attribute which has the highest accuracy score, then the \textit{Fairness Measure} $\mathcal{F}$ of a model is defined as the largest accuracy gap among all sensitive attributes computed as the minimum of the ratios of the accuracy scores of each sensitive attribute with respect to the dominant attribute.

{\small
\begin{equation}
    \mathcal{F}  = \min_{}(\frac{f(\mathbf{\hat{y}}, \mathbf{y}, s_0, \mathbf{x})}{f(\mathbf{\hat{y}}, \mathbf{y}, d, \mathbf{x})}, ..., \frac{f(\mathbf{\hat{y}}, \mathbf{y}, s_n, \mathbf{x})}{f(\mathbf{\hat{y}}, \mathbf{y}, d, \mathbf{x})})\\
\end{equation}
}In other words, $\mathcal{F}$ is defined as the ratio of the lowest accuracy for a sensitive attribute with respect to the highest accuracy value for that sensitive attribute.

\nl\textbf{\acf{CF}}:
\textit{Catastrophic forgetting}~\cite{kemker2018measuring} occurs when learning a new task negatively impacts previously learnt information. For our experiments, we also report the \textit{\ac{CF}} metric score~\cite{diaz2018don} for the \ac{CL} methods, measuring the average change in the accuracy scores of the \ac{CL} model for each previous task right after learning a new task.  This is computed as follows:

{\small
\[
    CF = \frac{\sum_{j=1}^{i-1} a_{j,j} - a_{i,j}}{i-1} \textrm{    ,    } 
    A= {\tiny \begin{bmatrix}
    a_{11} & 0  & \dots  & 0 \\
    a_{21} & a_{22}  & \dots  & 0 \\
    \vdots & \vdots  & \ddots & \vdots \\
    a_{n,1} & a_{n,2}  & \dots  & a_{n,n}
\end{bmatrix}
}\]}where $a_{i,j}$ denotes the accuracy of $i^{th}$ task right after learning $j^{th}$ task, $A$ is the matrix storing accuracy scores with dimensions $(n\times n)$ and $n$ is the number of classes.

\subsubsection{\textbf{Implementation Details}}
All models are trained using the \textit{adam} optimiser with a learning rate of $1.0e^{-4}$ and a batch-size of $24$. For the experiments with the RAF-DB dataset, all models are trained for $25$ epochs while for the BP4D dataset, due to a higher number of data samples, training converged after only $10$ epochs for all the approaches. All experiments are \textit{repeated} $3$ times and the results are \textit{averaged} across the repetitions to account for the random seeds. All models are implemented using the PyTorch Python Library based on the Continual Learning benchmarks provided by~\cite{Hsu18_EvalCL,van2019three}.

Table~\ref{tab:lambda_rafdb} and~\ref{tab:lambda_bp4d} report the regularisation coefficient values for the \ac{CL} methods for experiments with the RAF-DB and BP4D datasets, respectively. These values are set based on separate hyper-parameter searches for each model and selecting the best-performing values.

\section{Experiments and Results}
\label{sec:results}

\subsection{Experiment 1: Mitigating Bias in \acs{FER}}
\label{subsec:exp1}

In our experiments, we compare state-of-the-art bias mitigation approaches (see Section~\ref{subsec:traditionalbias}) with popular \ac{CL}-based methods (see Section~\ref{subsec:biascl}) on their ability to classify facial expressions without being affected by imbalances in data distributions.
Furthermore, to evaluate the applicability of \ac{CL} strategies as \textit{`fair'} \ac{FER} systems, we train and test these approaches on the RAF-DB dataset and compare their performance (both without and with data-augmentation) on learning to categorise $7$ expression classes, namely, \textit{surprise, sadness, happiness, fear, anger, disgust} and \textit{neutral}, with respect to $2$ different domain groups; gender (Male, Female) and race (Caucasian, African-American, Asian). 


\subsubsection{\textbf{Bias Across Gender Attributes}}

%

For the RAF-DB dataset, approximately $53.4\%$ of the samples are labelled as `Female' while the `Male' group constitutes about $40.3\%$ of the total samples. The rest of the samples are labelled as `Unsure' and omitted from our evaluations (see Fig.~\ref{fig:rafdb_gender}). As a result, the effective split of the dataset with respect to gender is somewhat balanced, $56.3\%$ Female against $43.7\%$ Male samples. For the non-\ac{CL}-based methods, the models are trained on the entire dataset and tested individually on the Male and Female subsets. For the \ac{CL} evaluations, however, the learning is split into two tasks corresponding to expression recognition for the Male (Task $1$) followed by expression recognition for the Female (Task $2$) sub-sets. Task-ordering is discussed further in Section~\ref{sec:discussfer}.

Table~\ref{tab:gender_raf_db_fairness} presents the experimental results comparing the different methods on their Accuracy as well as Fairness Measure scores. It can be seen that the \ac{CL} methods, overall, outperform all other methods both in accuracy as well as fairness scores while the baseline method performs the worst. Furthermore, although the accuracy scores of all the approaches increase when data-augmentation is used, not all of them are able to maintain fairness. \ac{CL} methods (with the exception of \ac{NR}) on the other hand, improve upon their fairness scores as well, with \ac{SI}~\cite{zenke2017continual} achieving the highest fairness scores both without and with data-augmentation.

\begin{table*}[t]
\centering
{
	\scriptsize
	\caption{\textbf{Experiment 1:} Race-wise Accuracy and Fairness Scores on RAF-DB dataset. Accuracy scores are reported after training the models on all the subsets. \textbf{Bold} values denote best while [\textit{bracketed}] denote second-best values for each column.}
	\label{tab:rafdb_race_fairness}
	\hspace*{-3mm}\begin{tabular}{c|c|c|c|c|c|c|c|c}\toprule
	
		\multirow{2}{*}{\textbf{Method}} & \multicolumn{3}{c|}{\textbf{Accuracy W/O Data-Augmentation}} & \multirow{2}{*}{\textbf{Fairness}} & \multicolumn{3}{c|}{\textbf{Accuracy W/ Data-Augmentation}} & \multirow{2}{*}{\textbf{Fairness}} \\ \cmidrule{2-4} \cmidrule{6-8}
 & \multicolumn{1}{c|}{\textit{\textbf{Caucasian}}} & \multicolumn{1}{c|}{\textit{\textbf{African American}}} &\multicolumn{1}{c|}{\textit{\textbf{Asian}}} & \multicolumn{1}{c|}{} & \multicolumn{1}{c|}{\textit{\textbf{Caucasian}}} & \multicolumn{1}{c|}{\textit{\textbf{African American}}} &\multicolumn{1}{c|}{\textit{\textbf{Asian}}} & \multicolumn{1}{c}{} \\ \midrule
 
	Baseline & 0.750$\pm$0.019 & 0.764$\pm$0.029 & \cellcolor{gray!25}\textbf{0.795$\pm$0.010} & 0.943 & 0.758$\pm$0.004 & 0.778$\pm$0.023 & \cellcolor{gray!25}\textbf{0.809$\pm$0.008} & 0.937 \\ 
	
	Offline  & 0.727$\pm$0.010 & 0.750$\pm$0.011 & 0.735$\pm$0.025 & 0.969 & 0.743$\pm$0.006 & 0.762$\pm$0.017 & 0.763$\pm$0.007 & 0.974\\ \midrule
	
	\multicolumn{9}{c}{\textbf{Non-\ac{CL}-based Bias Mitigation approaches}}\\ \midrule
	
	\ac{DDC}~\cite{Dwork2012Fairness} & 0.714$\pm$0.009 & 0.710$\pm$0.009 & 0.721$\pm$0.009 & 0.985 & 0.729$\pm$0.006 & 0.736$\pm$0.001 & 0.747$\pm$0.007 & 0.976 \\ 
	
	\ac{DIC}~\cite{wang2020towards} & 0.724$\pm$0.004 & 0.730$\pm$0.015 & 0.732$\pm$0.016 & 0.989 & 0.745$\pm$0.007 & 0.768$\pm$0.012 & 0.772$\pm$0.013 & 0.965 \\ 
	
	\ac{SS}~\cite{elkan2001foundations} & 0.734$\pm$0.005 & 0.728$\pm$0.015 & 0.757$\pm$0.014 & 0.961 & 0.748$\pm$0.002 & 0.752$\pm$0.019 & 0.767$\pm$0.023 & 0.975 \\ 
	
	\ac{DA}~\cite{xu2020investigating} & 0.634 & 0.584 & 0.544 & 0.858 & 0.756 & 0.766 & 0.704 & 0.919 \\ \midrule
	
	\multicolumn{9}{c}{\textbf{Continual Learning approaces}}\\ \midrule
	
	\ac{EWC}~\cite{kirkpatrick2017overcoming} & 0.764$\pm$0.011 & 0.758$\pm$0.002 & 0.768$\pm$0.016 & 0.987 & \cellcolor{gray!25}\textbf{0.796$\pm$0.006} & [\textit{0.788$\pm$0.009}] & 0.794$\pm$0.007 & 0.990 \\ 
	
	\ac{EWC}-Online~\cite{schwarz2018progress} & [\textit{0.773$\pm$0.018}] & 0.763$\pm$0.003 & 0.773$\pm$0.002 & 0.987 & 0.777$\pm$0.017 & 0.780$\pm$0.002 & 0.785$\pm$0.014 & 0.990 \\ 
	
	\ac{SI}~\cite{zenke2017continual} & 0.769$\pm$0.010 & [\textit{0.766$\pm$0.006}] & 0.769$\pm$0.008 & \cellcolor{gray!25}\textbf{0.996} & 0.785$\pm$0.013 & 0.783$\pm$0.003 & 0.782$\pm$0.009 & \cellcolor{gray!25}\textbf{0.996}\\ 
	
	\ac{MAS} \cite{aljundi2018memory} & 0.762$\pm$0.007 & 0.756$\pm$0.001 & 0.764$\pm$0.010 & [\textit{0.990}] & 0.781$\pm$0.017 & 0.776$\pm$0.012 & 0.781$\pm$0.007 & [\textit{0.994}] \\ 
	
	\ac{NR}~\cite{hsu2018re} & \cellcolor{gray!25}\textbf{0.779$\pm$0.015} & \cellcolor{gray!25}\textbf{0.772$\pm$0.017} & [\textit{0.793$\pm$0.002}] & 0.974 & [\textit{0.787$\pm$0.012}]& \cellcolor{gray!25}\textbf{0.796$\pm$0.005} & [\textit{0.808$\pm$0.014}] & 0.974 \\ \bottomrule
	
	\end{tabular}}
\end{table*}

\begin{table*}[t]
	\centering
	\scriptsize
	\caption{\textbf{Experiment 1:} \acf{CF} and Overall Accuracy (previous tasks) after each task for Race-ordered learning on RAF-DB dataset. \textbf{Bold} values denote best while [\textit{bracketed}] denote second-best values for each column.}
	\label{tab:rafdb_race_CF}
	\begin{tabular}{c|c|c|c|c|c|c|c|c|c|c|c|c}\toprule
		\multirow{3}{*}{\textbf{Method}} & \multicolumn{6}{c|}{\textbf{W/O Data-Augmentation}} & \multicolumn{6}{c|}{\textbf{W/ Data-Augmentation}} \\ \cmidrule{2-13} 
		
		 & \multicolumn{2}{c|}{Task 1} & \multicolumn{2}{c|}{Task 2} & \multicolumn{2}{c|}{Task 3} & \multicolumn{2}{c|}{Task 1} & \multicolumn{2}{c|}{Task 2} & \multicolumn{2}{c|}{Task 3} \\ \cmidrule{2-13} 
		 
		 & Acc. & CF & Acc. & CF & Acc. & CF & Acc. & CF & Acc. & \multicolumn{1}{c|}{CF} & \multicolumn{1}{c|}{Acc.} & \multicolumn{1}{c|}{CF} \\ \midrule
		 
		\ac{EWC}~\cite{kirkpatrick2017overcoming} & 0.777 & X & \cellcolor{gray!25}\textbf{0.754} & [\textit{0.025}] & 0.764 & 0.019 & 0.797 & X & \cellcolor{gray!25}\textbf{0.768} & \cellcolor{gray!25}\textbf{0.030} & \cellcolor{gray!25}\textbf{0.795} & [\textit{0.000}] \\ 
		
		\ac{EWC}-Online~\cite{schwarz2018progress} & [\textit{0.778}] & X & 0.738 & 0.046 & [\textit{0.779}] & 0.012 & 0.798 & X & [\textit{0.763}] & [\textit{0.038}] & 0.779 & 0.026 \\ 
		
		\ac{SI}~\cite{zenke2017continual} & \cellcolor{gray!25}\textbf{0.782} & X & 0.736 & 0.047 & 0.769 & \cellcolor{gray!25}\textbf{-0.003} & \cellcolor{gray!25}\textbf{0.802} & X & 0.759 & 0.044 & 0.784 & 0.007 \\ 
		
		\ac{MAS}~\cite{aljundi2018memory} & 0.766 & X & [\textit{0.744} & \cellcolor{gray!25}\textbf{0.023} & 0.762 & [\textit{0.003}] & [\textit{0.801}] & X & 0.762 & 0.039 & 0.780 & 0.004\\  
		
		\ac{NR}~\cite{hsu2018re} & [\textit{0.778}] & X & 0.734 & 0.049 & \cellcolor{gray!25}\textbf{0.781} & 0.009 & 0.784 & X & 0.744 & 0.043 & [\textit{0.790}] & \cellcolor{gray!25}\textbf{-0.006} \\ \bottomrule
		
	\end{tabular}
\end{table*}

In order to fully appreciate how \ac{CL} enables the models to retain their performance across the two tasks, it is important to understand how learning each new tasks impacts the model's performance on previously learnt tasks. Table~\ref{tab:gender_raf_db_cf} further reports the overall accuracy and \ac{CF} scores for the \ac{CL} methods after each task, evaluating the performance of these methods in terms of their ability to classify expressions for both male and female sub-sets. We observe that both with and without augmentation,  \ac{NR}~\cite{Hsu18_EvalCL} achieves the highest overall accuracy score after both tasks are learnt, while \ac{EWC} experiences the least forgetting. Furthermore, negative \ac{CF} scores for all \ac{CL} methods indicate that after learning task $2$, that is, to predict expressions on Female samples, the overall accuracy of the model increased for both Male and Female samples, without any forgetting occurring in the model.

\subsubsection{\textbf{Bias Across Race Attributes}}
The data distribution of the RAF-DB dataset is highly imbalanced with respect to Race with a majority ($77.4 \%$) of the samples labelled as Caucasian, while the African-American and Asian subsets correspond to only $7.1\%$ and $15.5\%$ of the samples, respectively (see Fig.~\ref{fig:rafdb_race}). Similar to the evaluations across gender, for the Non-\ac{CL}-based methods, the model is trained on the entire dataset and evaluated individually for the different race attributes. For the \ac{CL} evaluations, the learning is again split into three tasks corresponding to learning to predict expressions for Caucasian (Task $1$), African-American (Task $2$) and Asian (Task $3$) faces.

Table~\ref{tab:rafdb_race_fairness} presents the results of the experiments comparing Accuracy and Fairness Measure scores. The imbalances in the data distribution affect all the approaches such that the model accuracy varies across the race groupings. Yet, the \ac{CL} methods seem to handle this best, achieving comparable accuracy with high fairness scores. Even though the \ac{CL} approaches are not always the best performing ones, particularly for the \textit{Asian} subset, all of them achieve high fairness scores, with \ac{SI} performing the best. This underlines their ability to balance learning across the different tasks. They are able to give preference to being consistent and \textit{fair}, trading-off higher accuracy scores for any individual race label. The \ac{NR} approach, on the other hand, achieves the highest accuracy scores on Task $1$ and Task $2$, owing to the explicit replay mechanism, but sacrifices fairness across all groups in the process. Additionally, as RAF-DB is a relatively small dataset, data-augmentation has a positive effect on accuracy scores of all the models but the fairness scores do not change significantly. 

In Table~\ref{tab:rafdb_race_CF}, the accuracy and \ac{CF} scores can be seen for all the \ac{CL} methods reporting model performance on all previous tasks computed at the end of each new task. We see that all models tend to forget as they learn new tasks, yet the \ac{SI} method is able to mitigate forgetting the best after having learnt all the tasks. When data-augmentation is used, the individual accuracy scores are enhanced but the \ac{CF} scores do not improve. Owing to the explicit replay mechanism, the \ac{NR} method performs the best in terms of mitigating forgetting when using data-augmentation.

\begin{table*}[t]
	\centering
	{
	\scriptsize
	\caption{\textbf{Experiment 2:} Gender-wise Accuracy and Fairness Scores on BP4D dataset. Accuracy scores are reported after training the models on both Male and Female subsets. \textbf{Bold} values denote best while [\textit{bracketed}] denote second-best values for each column.} 
	\label{tab:bp4d_gender_fairness}
	\vspace{-2mm}
	\begin{tabular}{c|c|c|c|c|c|c}\toprule
		\multirow{2}{*}{\textbf{Method}} & \multicolumn{2}{c|}{\textbf{Accuracy W/O Data-Augmentation}} & \multicolumn{1}{c|}{\multirow{2}{*}{\textbf{Fairness}}} & \multicolumn{2}{c|}{\textbf{Accuracy W/ Data-Augmentation}} & \multicolumn{1}{c}{\multirow{2}{*}{\textbf{Fairness}}} \\ \cmidrule{2-3} \cmidrule{5-6}
		
		 & \multicolumn{1}{c|}{\textit{\textbf{Male}}} & \multicolumn{1}{c|}{\textit{\textbf{Female}}} & \multicolumn{1}{c|}{} & \multicolumn{1}{c|}{\textit{\textbf{Male}}} & \multicolumn{1}{c|}{\textit{\textbf{Female}}} & \multicolumn{1}{c}{} \\ \midrule
		 
		Baseline & 0.691$\pm$0.007 & 0.718$\pm$0.004 &  0.962 & 0.692$\pm$0.009 & 0.735$\pm$0.004 &  0.941 \\ 
		
		Offline  & 0.701$\pm$0.007 & 0.712$\pm$0.006 & 0.984 & 0.731$\pm$0.005 & 0.735$\pm$0.008 &  [\textit{0.994}]\\ \midrule
		
		\multicolumn{7}{c}{\textbf{Non-\ac{CL}-based Bias Mitigation approaches}}\\ \midrule

		\ac{DDC}~\cite{Dwork2012Fairness}& 0.740$\pm$0.008 & 0.733$\pm$0.005 & [\textit{0.990}]& 0.752$\pm$0013 & 0.745$\pm$0.007 &  0.991 \\ 
		
		\ac{DIC}~\cite{wang2020towards} & 0.734$\pm$0.010 & 0.750$\pm$0.008 & 0.979 & 0.741$\pm$0.012 & 0.752$\pm$0.009 &  0.985\\ 
		
		\ac{SS}~\cite{elkan2001foundations} & 0.715$\pm$0.003 & 0.732$\pm$0.002 & 0.977 & 0.735$\pm$0.005 & 0.748$\pm$0.005 &  0.983\\ 
		
		\ac{DA}~\cite{xu2020investigating} & 0.731 & 0.735 &  \cellcolor{gray!25}\textbf{0.994} & 0.776 & \cellcolor{gray!25}\textbf{0.772} &  \cellcolor{gray!25}\textbf{0.995} \\ \midrule
		
		\multicolumn{7}{c}{\textbf{Continual Learning approaches}}\\ \midrule
		
		\ac{EWC}~\cite{kirkpatrick2017overcoming}  &[\textit{0.769$\pm$0.006}] &  0.754$\pm$ 0.008 &  0.981 &0.772$\pm$0.006 &  0.766$\pm$0.006 & 0.992 \\ 
		
		\ac{EWC}-Online~\cite{schwarz2018progress}&0.762$\pm$0.009 & 0.744$\pm$ 0.006 & 0.976 &  0.772$\pm$0.012 &[\textit{0.767$\pm$0.004}] & [\textit{0.994}] \\ 
		
		\ac{SI}~\cite{zenke2017continual} & 0.755$\pm$0.021 & 0.745$\pm$ 0.016 &  0.986 & [\textit{0.778$\pm$0.017}] & 0.751$\pm$0.022 &  0.965 \\ 
		
		\ac{MAS} \cite{aljundi2018memory} & \cellcolor{gray!25}\textbf{0.788$\pm$0.007} &  \cellcolor{gray!25}\textbf{0.761$\pm$ 0.009} & 0.966 & \cellcolor{gray!25}\textbf{0.784$\pm$0.003} & 0.758$\pm$0.012 &  0.967  \\ 
		
		\ac{NR}~\cite{hsu2018re}& [\textit{0.769$\pm$0.010}] &  [\textit{0.756$\pm$ 0.004}] & 0.983  & 0.774$\pm$0.009  &0.739$\pm$0.016& 0.954 \\ \bottomrule
	
	\end{tabular}}
	\vspace{-3mm}
\end{table*}

\begin{table*}[t]
	\centering
	{
	\scriptsize
	\caption{\textbf{Experiment 2:} CF and Overall Accuracy (previous tasks) after each task for Gender-ordered learning on BP4D dataset. \textbf{Bold} values denote the best while [\textit{bracketed}] denote second-best values for each column.}
	\label{tab:bp4d_gender_CF}
	\vspace{-2mm}
	\begin{tabular}{c|cc|cc|cc|cc}\toprule
		
		\multirow{3}{*}{\textbf{Method}} & \multicolumn{4}{c|}{\textbf{W/O Data-Augmentation}} & \multicolumn{4}{c}{\textbf{W/ Data-Augmentation}} \\ \cmidrule{2-9} 
		
		 & \multicolumn{2}{c|}{Task 1} & \multicolumn{2}{c|}{Task 2} & \multicolumn{2}{c|}{Task 1} & \multicolumn{2}{c}{Task 2} \\ \cmidrule{2-9} 
		 
		 & Acc. & CF & Acc. & CF & Acc. & CF & Acc. & CF \\ \midrule
		 
		\ac{EWC}~\cite{kirkpatrick2017overcoming} & [\textit{0.745}] & X & [\textit{0.768}] & [\textit{-0.020}] & 0.729 & X & [\textit{0.761}] & \cellcolor{gray!25}\textbf{-0.024} \\ 
		
		\ac{EWC} Online~\cite{schwarz2018progress} & 0.744 & X & \cellcolor{gray!25}\textbf{0.769} & \cellcolor{gray!25}\textbf{-0.022} & 0.728 & X & 0.753 & [\textit{-0.016}] \\ 
		
		\ac{SI}~\cite{zenke2017continual} & \cellcolor{gray!25}\textbf{0.753} & X & 0.764 & 0.002 & [\textit{0.736}] & X & 0.749 & -0.008 \\ 
		
		\ac{MAS}~\cite{aljundi2018memory} & 0.734 & X & 0.765 & -0.011 & \cellcolor{gray!25}\textbf{0.745} & X & \cellcolor{gray!25}\textbf{0.766} & -0.002 \\ 
		
		\ac{NR}~\cite{hsu2018re} & 0.740 & X & 0.759 & 0.000 & [\textit{0.736}] & X & 0.758 & -0.006 \\ \bottomrule
		
\end{tabular}}
\vspace{-4mm}
\end{table*}

\subsection{Experiment 2: Mitigating Bias in \acs{AU} Detection}
\label{subsec:exp2}
As more than one \ac{AU} may be \textit{activated} at the same time (for example, \ac{AU} $1,2$ and $26$ together may depict \textit{surprise}), predicting facial \acsp{AU} poses a multi-label classification problem. Imbalances in data distributions with respect to the different \textit{gender} and \textit{race} attributes become even more prominent with certain \ac{AU} classes having much more data samples than the others (see Fig.~\ref{fig:bp4d_distribution}). We compare different bias mitigation strategies (see Section~\ref{subsec:traditionalbias}) with \ac{CL}-based methods (see Section~\ref{subsec:biascl}) to understand how they cope with imbalances in data distributions while retaining model performance. 

Similar to Experiment~$1$, we compare the performance of all models (without and with data-augmentation) on detecting activations for the $12$ \acsp{AU} for \textit{gender} (Male, Female) and \textit{race} (White, Black, Asian, Latino) groupings.


\subsubsection{\textbf{Bias Across Gender Attributes}}





Similar to Experiment~$1$, for the non-\ac{CL}-based methods, we train the individual models on the entire dataset but evaluate them individually for Male and Female subsets. The BP4D data distribution is skewed in favour of Female samples constituting $60.96\%$ of the data while only $39.04\%$ samples belong to the Male sub-set (see Fig.~\ref{fig:bp4d_gender}). For \ac{CL} methods, the learning is split into two \textit{tasks}: Task $1$: Male and Task $2$: Female, incrementally learning with the two data splits. Task-ordering and its affect is discussed in Section~\ref{sec:discussau}. 

Table~\ref{tab:bp4d_gender_fairness} compares the different methods on their Accuracy and Fairness Measure scores for both Male and Female subsets. \ac{CL} methods are shown to outperform other methods in terms of accuracy, yet, \ac{DA}~\cite{xu2020investigating} attains the highest Fairness score. Although \ac{CL} methods perform better than others on individual tasks, they are not able to balance this learning across tasks as compared to \ac{DA}. The \ac{MAS} outperforms other \ac{CL} methods on accuracy, yet \ac{SI} and \ac{EWC}-Online methods score higher on fairness, without and with data-augmentation, respectively. Data-augmentation, overall, has a positive impact on model accuracy scores but much like Experiment~$1$, it does not impact model fairness, significantly. Individual class accuracy between Male and Female splits does not vary significantly for the $12$ \ac{AU} labels with \ac{AU} $2$ and \ac{AU} $12$ achieving the lowest and highest accuracy scores, respectively, across the models for both the splits. This can be due to these classes consisting of the lowest and highest number of data samples in BP4D distribution across both gender and race splits (see Fig~\ref{fig:bp4d_distribution}).

Comparing different \ac{CL} methods on their ability to maintain performance across the tasks, Table~\ref{tab:bp4d_gender_CF} reports the overall accuracy and \ac{CF} scores for the models. Owing to the complex multi-label nature of the tasks as well as the high gender disparity in the data distribution, we see a high variation in performance scores of the different \ac{CL} methods. While \ac{EWC}-Online performs the best without employing data-augmentation achieving a negative \ac{CF} score, the \ac{MAS} model performs the best with data-augmentation. 

\begin{table*}[t]
	\centering
	\setlength\tabcolsep{3.0pt}

	{
	\scriptsize

	\caption{\textbf{Experiment 2:} Race-wise Accuracy and Fairness Scores on BP4D dataset. Accuracy scores are reported after training the models on all the subsets. \textbf{Bold} values denote best while [\textit{bracketed}] denote second-best values for each column.}
	\label{tab:bp4d_race_fairness}
	\begin{tabular}{c|c|c|c|c|c|c|c|c|c|c}\toprule
		\multirow{2}{*}{\textbf{Method}} & \multicolumn{4}{c|}{\textbf{Accuracy W/O Data-Augmentation}} & \multirow{2}{*}{\textbf{Fairness}} & \multicolumn{4}{c|}{\textbf{Accuracy W/ Data-Augmentation}} & \multirow{2}{*}{\textbf{Fairness}} \\ \cmidrule{2-5} \cmidrule{7-10}
		
		 & \multicolumn{1}{c|}{\textit{\textbf{Black}}} & \multicolumn{1}{c|}{\textit{\textbf{Asian}}} &\multicolumn{1}{c|}{\textit{\textbf{White}}}&\multicolumn{1}{c|}{\textit{\textbf{Latino}}} & \multicolumn{1}{c|}{} & \multicolumn{1}{c|}{\textit{\textbf{Black}}} & \multicolumn{1}{c|}{\textit{\textbf{Asian}}} &\multicolumn{1}{c|}{\textit{\textbf{White}}}&\multicolumn{1}{c|}{\textit{\textbf{Latino}}} & \multicolumn{1}{c}{} \\ \midrule
		 
		Baseline &0.659$\pm$0.048 & 0.720$\pm$0.016 & 0.771$\pm$0.018 & 0.764$\pm$0.008 & 0.855 & 0.654$\pm$0.042 & 0.685$\pm$0.022& 0.763$\pm$0.009& 0.737$\pm$0.019 & 0.858 \\ 
		
		Offline &0.694$\pm$0.009  &0.724$\pm$0.019  &[\textit{0.781$\pm$0.012}]  & \cellcolor{gray!25}\textbf{0.790$\pm$0.011}  &0.878  &0.706$\pm$0.008  &0.750$\pm$0.017  & 0.783$\pm$0.011  &0.777$\pm$0.019  & 0.901 \\ \midrule
		
		\multicolumn{11}{c}{\textbf{Non-\ac{CL}-based Bias Mitigation Approaches}} \\ \midrule
		
		\ac{DDC}~\cite{Dwork2012Fairness} &0.722$\pm$0.018  &0.729$\pm$0.007  &0.775$\pm$0.015 &[\textit{0.785$\pm$0.005}]  &0.920  &0.727$\pm$0.004  &0.731$\pm$0.013  &0.781$\pm$0.010  &[\textit{0.787$\pm$0.005}]  &0.924  \\ 
		
		\ac{DIC}~\cite{wang2020towards} &0.712$\pm$0.011  &0.731$\pm$0.014  &0.767$\pm$0.017  &0.770$\pm$0.015  &0.925  &0.719$\pm$0.009  &0.742$\pm$0.009  &0.778$\pm$0.005  &0.780$\pm$0.012  &0.922  \\ 
		
		\ac{SS}~\cite{elkan2001foundations} &0.722$\pm$0.007 &0.729$\pm$0.012  &0.775$\pm$0.012  &[\textit{0.785$\pm$0.013}]  &0.920  &0.723$\pm$0.018  &0.731$\pm$0.007  &0.781$\pm$0.002  &[\textit{0.787$\pm$0.006}]  &0.919 \\ 
		
		\ac{DA}~\cite{xu2020investigating} &0.706 &0.736  &0.740  &0.736  &[\textit{0.954}]  & \cellcolor{gray!25}\textbf{0.767}  &0.774  &0.771  & \cellcolor{gray!25}\textbf{0.797}  & [\textit{0.962}] \\ \midrule
		
		\multicolumn{11}{c}{\textbf{Continual Learning Approaches}} \\ \midrule
		
		\ac{EWC}~\cite{kirkpatrick2017overcoming} &[\textit{0.742$\pm$0.012}] & 0.760$\pm$0.016 & 0.769$\pm$0.007 & 0.731$\pm$0.007 & 0.949& 0.721$\pm$0.035 & 0.754$\pm$0.019&0.764$\pm$0.003 & 0.738$\pm$0.009 & 0.943 \\ 

		\ac{EWC}-Online~\cite{schwarz2018progress} &[\textit{0.742$\pm$0.004}] & 0.769$\pm$0.024 & 0.762$\pm$0.008 & 0.720$\pm$0.000 & 0.937& 0.748$\pm$0.004 & 0.776$\pm$0.008& 0.781$\pm$0.009& 0.747$\pm$0.006 & 0.957\\ 
		
		\ac{SI}~\cite{zenke2017continual} &0.731$\pm$0.024 & [\textit{0.773$\pm$0.000}] &0.762$\pm$0.000  & 0.770$\pm$0.007 & 0.946& 0.752$\pm$0.018  & \cellcolor{gray!25}\textbf{0.788$\pm$0.004}& [\textit{0.782$\pm$0.008}]& 0.765$\pm$0.001 & 0.954 \\ 
		
		\ac{MAS}~\cite{aljundi2018memory}  & 0.710$\pm$0.023 & [\textit{0.773$\pm$0.006}] &0.753$\pm$0.009 & 0.772$\pm$0.003 & 0.920 & 0.715$\pm$0.021   & [\textit{0.786$\pm$0.008}]& 0.763$\pm$0.009& 0.775$\pm$0.012 & 0.909  \\ 
		
		\ac{NR}~\cite{hsu2018re} & \cellcolor{gray!25}\textbf{0.767$\pm$0.023} & \cellcolor{gray!25}\textbf{0.779$\pm$0.009} & \cellcolor{gray!25}\textbf{0.788$\pm$0.001} & 0.762$\pm$0.012 & \cellcolor{gray!25}\textbf{0.966} & [\textit{0.763$\pm$0.006}] & 0.780$\pm$0.005&  \cellcolor{gray!25}\textbf{0.783$\pm$0.012}& 0.764$\pm$0.002 & \cellcolor{gray!25}\textbf{0.974}  \\ \bottomrule
		
	\end{tabular}}
\end{table*}

\begin{table*}[t]
	\centering
	\setlength\tabcolsep{4.5pt}

	{
	\scriptsize
	\caption{\textbf{Experiment 2:} CF and Overall Accuracy (previous tasks) after each task for Race-ordered learning on BP4D dataset. \textbf{Bold} values denote the best while [\textit{bracketed}] denote second-best values for each column.}
	\label{tab:bp4d_race_CF}
	\begin{tabular}{c|cc|cc|cc|cc|cc|cc|cc|cc}\toprule
		\multirow{3}{*}{\textbf{Method}} & \multicolumn{8}{c|}{\textbf{W/O Data-Augmentation}} & \multicolumn{8}{c}{\textbf{W/ Data-Augmentation}} \\ \cmidrule{2-17} 
		
		 & \multicolumn{2}{c|}{Task 1} & \multicolumn{2}{c|}{Task 2} & \multicolumn{2}{c|}{Task 3} & \multicolumn{2}{c|}{Task 4} & \multicolumn{2}{c|}{Task 1} & \multicolumn{2}{c|}{Task 2} & \multicolumn{2}{c|}{Task 3} &  \multicolumn{2}{c}{Task 4}  \\ \cmidrule{2-17}  
		 
		 & Acc. & CF & Acc. & CF & Acc. & CF & Acc. & CF & Acc. & CF & Acc. & CF & Acc. & CF & Acc. & CF \\ \midrule
		 
		\ac{EWC}~\cite{kirkpatrick2017overcoming}  & 0.763 & X & 0.707 & 0.059 & 0.692 & [\textit{0.125}] & 0.761 & [\textit{-0.042}] & 0.763 & X & 0.686 & 0.087 & 0.673 & \cellcolor{gray!25}\textbf{0.129} & 0.748 & -0.026 \\ 
		
		\ac{EWC}-Online~\cite{schwarz2018progress} & \cellcolor{gray!25}\textbf{0.773} & X & 0.693 & 0.079 & 0.682 & \cellcolor{gray!25}\textbf{0.122} & 0.757 & -0.032 & 0.766 & X & 0.697 & 0.081 & 0.682 & [\textit{0.140}] & 0.754 & -0.026 \\ 
		
		\ac{SI}~\cite{zenke2017continual} & [\textit{0.765}] & X & \cellcolor{gray!25}\textbf{0.734} & \cellcolor{gray!25}\textbf{0.044} & \cellcolor{gray!25}\textbf{0.714} & 0.154 & [\textit{0.764}] & -0.027 & [\textit{0.774}] & X & 0.702 & 0.091 & [\textit{0.686}] & 0.147 & [\textit{0.768}] & -0.032 \\ 
		
		\ac{MAS} \cite{aljundi2018memory}  & 0.759 & X & [\textit{0.724}] & [\textit{0.048}] & [\textit{0.707}] & 0.144 & 0.762 & -0.035 & 0.754 & X & \cellcolor{gray!25}\textbf{0.717} & \cellcolor{gray!25}\textbf{0.056} & \cellcolor{gray!25}\textbf{0.697} & 0.162 & 0.762 & [\textit{-0.033}] \\ 
		
		\ac{NR}~\cite{hsu2018re} & 0.759 & X & 0.710 & 0.055 & 0.690 & 0.143 & \cellcolor{gray!25}\textbf{0.777} & \cellcolor{gray!25}\textbf{-0.053} & \cellcolor{gray!25}\textbf{0.775} & X & [\textit{0.705}] & [\textit{0.079}] & 0.685 & 0.156 & \cellcolor{gray!25}\textbf{0.780} & \cellcolor{gray!25}\textbf{-0.035} \\ \bottomrule
		
	\end{tabular}}
\end{table*}

\subsubsection{\textbf{Bias Across Race Attributes}}
The majority of the samples in the BP4D dataset are labelled as White (approximately $46.76\%$) with other samples corresponding to Asian ($26.08\%$), Black ($16.56\%$) and Latino ($10.6\%$) groups (see Fig.~\ref{fig:bp4d_race}). For our evaluations, we split the dataset into $4$ subsets based on these labels, representing the $4$ tasks, that is, Task $1$: Black, Task $2$: Asian, Task $3$: White and Task $4$: Latino, for the \ac{CL} models. Task-orderings are discussed further in Section~\ref{sec:discussau}.

It can be seen in Table~\ref{tab:bp4d_race_fairness} that \ac{CL} methods, overall, achieve high accuracy and Fairness scores, with the \ac{NR} approach outperforming other methods both without and with data-augmentation. While \ac{NR} achieves the highest accuracy scores for Black, Asian and White subsets even without data-augmentation, none of the approaches is able to beat the off-line baseline for the Latino subset. This may be owing to the extremely low sample-rate for the Latino subset that the \ac{CL} methods, instead of focusing on improving performance on this sub-set, focus on maintaining performance across all the sub-sets. Data-augmentation has a positive impact on all the models in terms of improvements both in accuracy and fairness scores.

Different \ac{CL} models handle the high variance in data distribution with respect to racial identity labels with varying levels of success. Table~\ref{tab:bp4d_race_CF} shows how, at different points during the learning, different models perform better than others, while \ac{NR} achieves the highest accuracy and \ac{CF} scores after all tasks are learnt, both with and without data-augmentation. The negative \ac{CF} scores for all the approaches at the end of Task $4$ signifies that all the \ac{CL} models were able to mitigate forgetting and the overall model performance improved as they incrementally learnt new tasks.


\section{Discussion}
\label{sec:discussion}

Our experiments on \ac{FER} (see Section~\ref{subsec:exp1}) and \ac{AU} detection (see Section~\ref{subsec:exp2}) tasks are motivating as they highlight how adopting \ac{CL} strategies may enable \textit{fairer} facial affect analysis algorithms. Consistently achieving high accuracy as well as fairness measure scores, \ac{CL} offers an improvement over existing learning strategies for bias mitigation in \ac{ML} algorithms. Robustly managing imbalances in data distributions, both without and with data-augmentation, \ac{CL} methods are better equipped to deal with biases owing to their learning strategy of focusing on one domain group at a time. Here, we discuss each task individually and highlight how \ac{CL} provides a solution towards \textit{fairer} facial affect analyses.

\subsection{Facial Expression Recognition}
\label{sec:discussfer}

When applied to \ac{FER}, \ac{CL} methods aim to sequentially learn to predict expression categories for the different gender and race groups. The models are trained with one group at a time and as the model experiences samples from other groups, it is actively trying to maintain performance at previously seen groups without forgetting. As a result, for both gender and race groups, \ac{CL} models are able to achieve high fairness scores by balancing performance across the splits, with the \ac{SI} model performing the best (see Table~\ref{tab:rafdb_fair}). Selective updates of network parameters in order to mitigate forgetting allows \ac{CL} models to maintain high accuracy scores across the different gender and race attributes. This makes them distinct from other approaches, directly focusing on maintaining performance across different domain distributions instead of deciding whether to capture domain-specific features or not.  In comparison, non-\ac{CL}-based methods rely on becoming `aware' of domain attributes to predict expressions according to the subjects sharing gender or race attributes or learning feature representations that actively `block' domain discriminative features~\cite{xu2020investigating}. Furthermore, for most of the non-\ac{CL} methods, with the exception of \ac{DA}, we need to know the domain groupings a priori which may not always be possible in real-world scenarios. For \ac{CL} methods, however, as models learn sequentially, there is no need to provide any domain information a priori and learning can be extended to new domains.

One concern when applying \ac{CL} methods to \ac{FER} tasks is the class-ordering effect where model performance is seen to be sensitive to the order in which it learns different expression classes~\cite{churamani2020clifer}. In our experiments, as we implement the \ac{Domain-IL} scenario where all classes are learnt at the same time, albeit one domain-group at a time, class-ordering does not play any role in the learning. Instead, we explore whether different task-orderings, that is, learning with different sequences of gender or race group splits has any effect on the models' ability to maintain performance. For both gender and race domains, we experiment with different orders of learning the tasks but no significant effect of domain ordering is witnessed on the models' performance. Class-wise accuracies are largely consistent between the different learning settings for the \ac{CL} models with model accuracy being the worst for \textit{disgust} due to the overall low number of samples, in line with what was reported in~\cite{xu2020investigating}.

\begin{table}[t]
	\centering
	\setlength\tabcolsep{3.0pt}

	{
	\scriptsize

	\caption{\textbf{Experiment 1:} Fairness Measure Scores across Gender and Race distributions for the RAF-DB Dataset. \textbf{Bold} values denote best while [\textit{bracketed}] denote second-best values for each column.}
	\label{tab:rafdb_fair}
	\begin{tabular}{c|cc|cc}\toprule
		\multirow{2}{*}{\textbf{Method}} & \multicolumn{2}{c|}{\textbf{W/O Data-Augmentation}} & \multicolumn{2}{c}{\textbf{W/ Data-Augmentation}} \\ \cmidrule{2-5} 
		
		 & \textit{\textbf{Gender}} & \textit{\textbf{Race}} 
		 & \textit{\textbf{Gender}} & \textit{\textbf{Race}} 
		 \\ \midrule
		Baseline & 0.834 & 0.943 
		& 0.816 & 0.937  
		\\ 
		
		Offline Training & 0.944 & 0.925 
		& 0.954 & 0.974 
		\\ \midrule

		\multicolumn{5}{c}{\textbf{Non-\ac{CL}-based Bias Mitigation Methods}} \\ \midrule

		\ac{DDC}~\cite{wang2020towards} & 0.968 &  0.985  
		& 0.961 & 0.976  
		\\ 

		\ac{DIC}~\cite{wang2020towards} & 0.938 & 0.989 
		& 0.962 & 0.965 
		\\ 
		\ac{SS}~\cite{elkan2001foundations} & 0.955 & 0.961 
		& 0.954 & 0.975 
		\\ 

		\ac{DA}~\cite{xu2020investigating} & 0.975 & 0.858 
		& [\textit{0.997}] & 0.919 
		\\ \midrule

		\multicolumn{5}{c}{\textbf{Continual Learning Methods}} \\ \midrule
		\ac{EWC}~\cite{kirkpatrick2017overcoming} & 0.972 & 0.987 
		& 0.983 & 0.990 
		\\ 
		\ac{EWC}-Online~\cite{schwarz2018progress} & 0.970 & 0.987 
		& 0.974 & 0.990 
		\\ 
		\ac{SI}~\cite{zenke2017continual} & \cellcolor{gray!25}\textbf{0.990} & \cellcolor{gray!25}\textbf{0.996} 
		& \cellcolor{gray!25}\textbf{0.999} & \cellcolor{gray!25}\textbf{0.996} 
		\\ 
		\ac{MAS}~\cite{aljundi2018memory} & [\textit{0.980}] & [\textit{0.990}] 
		& 0.990 & [\textit{0.994}] 
		\\ 
		\acs{NR}~\cite{hsu2018re} &0.928 & 0.974 
		& 0.923 & 0.974 
		\\ \bottomrule
\end{tabular}}
\end{table}
\subsection{Action Unit Detection}
\label{sec:discussau}
\acf{AU} detection poses a \textit{harder} multi-label classification problem where the models need to predict all the \acsp{AU} activated in a given sample. The inherent class-imbalances in the BP4D dataset are further accentuated by the imbalances with respect to gender and race attributes, making it extremely difficult for \ac{CL} as well as non-\ac{CL} models to maintain performance across the different groups. The under-represented classes are reduced to even fewer samples per class when split across gender or race, making it even more difficult for these models to cope with data imbalances. Even though \ac{CL}-based methods are able to achieve the highest individual accuracy scores (averaged across the $12$ \acsp{AU}) for most of the gender and race groups, this comes at the cost of balancing learning across the different attributes. For the gender splits, \acf{DA} achieves the highest fairness scores, despite performing moderately in terms of accuracy on individual splits (see Table~\ref{tab:bp4d_fairness}). Blocking individual domain-specific information, allows \ac{DA} to balance learning across the different splits, resulting in high fairness scores. For \ac{CL} models, however, the multi-label classification settings cause the models to focus more on overall individual performance rather than on maintaining performance across the gender splits. In the case of race-splits, we see that the \ac{NR} approach achieves the highest fairness scores, while \ac{DA} performs second-best. This is due to the memory-intensive rehearsal mechanism that physically stores and replays samples from previously seen domains to retain model performance. Even though regularisation-based approaches target accuracy and trade-off fairness in the process, they still perform better than most non-\ac{CL}-based methods. 

Due to the multi-label settings, all classes are learnt together with no ordering of the classes required. Furthermore, domain-ordering, that is, in which order the gender and race domains should be learnt, does not have any significant effect on model performance for the \ac{CL} methods. Owing to the highly imbalanced class-distributions, the performance of all models are poor for under-represented classes such as \ac{AU} $1, 2$ and $4$, across all gender and race splits. On the other hand, the highest model performances are achieved for dominant classes such as \acsp{AU} $10$ and $12$. These results are inline with other \ac{AU} prediction approaches~\cite{Churamani2020AULACaps, li2019semantic, shao2020spatiotemporal} that report similar differences in performance across these \acsp{AU}. 


\begin{table}[t]
	\centering
	\setlength\tabcolsep{3.0pt}

	{
	\scriptsize
	\caption{\textbf{Experiment 2:} Fairness Measure Scores across Gender and Race distributions for the BP4D Dataset. \textbf{Bold} values denote best while [\textit{bracketed}] denote second-best values for each column.}
	\label{tab:bp4d_fairness}
	\begin{tabular}{c|cc|cc}\toprule
	
		\multirow{2}{*}{\textbf{Method}} & \multicolumn{2}{c|}{\textbf{W/O Data-Augmentation}} & \multicolumn{2}{c}{\textbf{W/ Data-Augmentation}} \\ \cmidrule{2-5} 
		
		 & \textit{\textbf{Gender}} & \textit{\textbf{Race}} & \textit{\textbf{Gender}} & \textit{\textbf{Race}} \\ \midrule
		 
		Baseline & 0.962 & 0.855 & 0.941 & 0.858 \\
		
		Offline & 0.984 & 0.878 & [\textit{0.994}] & 0.901 \\ \midrule
		
		\multicolumn{5}{c}{\textbf{Non-\ac{CL}-based Bias Mitigation Approaches}} \\ \midrule
		
		\ac{DDC}~\cite{wang2020towards} & [\textit{0.990}] & 0.920 & 0.991 & 0.924 \\ 
		
		\ac{DIC}~\cite{wang2020towards} & 0.979 & 0.925 & 0.985 & 0.922 \\ 
		
		\ac{SS}~\cite{elkan2001foundations} & 0.977 & 0.920 & 0.983 & 0.919 \\
		
		\ac{DA}~\cite{xu2020investigating} & \cellcolor{gray!25}\textbf{0.994} & [\textit{0.954}] & \cellcolor{gray!25}\textbf{0.995} & [\textit{0.962}] \\ \midrule
		
		\multicolumn{5}{c}{\textbf{Continual Learning Approaches}} \\ \midrule
		
		\ac{EWC}~\cite{kirkpatrick2017overcoming} & 0.981 & 0.949 & 0.992 & 0.943 \\ 
		
		\ac{EWC}-Online~\cite{schwarz2018progress} & 0.976 & 0.937 & [\textit{0.994}] & 0.957 \\ 
		
		\ac{SI}~\cite{zenke2017continual} & 0.986 & 0.946 & 0.965 & 0.954 \\ 

		\ac{MAS}~\cite{aljundi2018memory} & 0.966 & 0.920 & 0.967 & 0.909 \\ 
		
		\ac{NR}~\cite{hsu2018re} & 0.983 & \cellcolor{gray!25}\textbf{0.966} & 0.954 & \cellcolor{gray!25}\textbf{0.974} \\ \bottomrule
	\end{tabular}}
\end{table}

\subsection{Limitations of \ac{CL}-based Bias Mitigation}
Our benchmark experiments with the RAF-DB and BP4D datasets highlight the potentials of  \ac{CL}-based models for creating \textit{fairer} facial expression recognition  systems. \ac{CL}-based models outperform other bias mitigation strategies for evaluations across gender and race domains, managing shifts in data distributions well. However, more work is needed to optimise \ac{CL}-based models for multi-label settings where they under-perform (see Table~\ref{tab:bp4d_gender_fairness}~and~\ref{tab:bp4d_race_fairness}). Recent work by \et{Kim}~\cite{kim-ECCV-2020} proposes a new replay-based strategy, the \acf{PRS}, that aims to tackle continual learning for multi-label classification, balancing both intra- and inter-task imbalances. Yet, they benchmark their approach on classification settings with little-to-no overlap between the tasks. This is not the case for \ac{AU} detection where the different domains, as well as the classes within each domain, share feature representations, making it even harder for the models.  

Furthermore, as regularisation-based \ac{CL} models assign \textit{importance} to different parameters based on their contribution towards previously learnt tasks, shared feature representations makes it harder for models to incrementally learn different tasks/domains as model parameters may contribute to more than one task or domain. Rehearsal-based methods such as \ac{NR}, on the other hand, require the models to physically store seen samples from previous tasks, interleaving them with new data to maintain performance. As the number of tasks, or in the case of \ac{Domain-IL}, data-splits across domains such as gender or race increase, storing samples from all the domains becomes extremely expensive both in terms of its memory footprint as well as the computational power needed to train the algorithms. 

Additionally, as the tasks increase, models may experience saturation~\cite{Parisi2018a} requiring stronger regularisation in the models to be able to preserve past knowledge~\cite{Titsias2020Functional}. The performance of the models also takes a hit where the model needs to re-prioritise whether to give more importance to the new task or remembering previous tasks. We see this in race-wise splits for both the datasets (see Table~\ref{tab:rafdb_race_fairness}~and~\ref{tab:bp4d_race_fairness}) where regularisation-based models attain higher accuracy scores for the last split, while the \ac{NR} method aims to maintain a higher fairness score instead. 

\section{Conclusion and Future Work}
\label{sec:conclusion}

In this work, we propose the novel use of Domain Incremental \ac{CL} as a potent bias mitigation method for facial analysis tasks. In particular, we highlight how using \ac{Domain-IL} settings, regularisation-based \ac{CL} methods can help develop \textit{fairer} expression recognition and \ac{AU} detection algorithms. Our experiments with popular benchmark datasets, RAF-DB for expression recognition and BP4D for \ac{AU} detection, showcase the superlative performance of \ac{CL} methods at handling imbalances in data distributions with respect to demographic attributes of gender and race. In comparison with state-of-the-art bias mitigation approaches, these methods are able to balance learning across different domain splits, not only achieving high accuracy scores but also maintaining fairness across the different splits. 

Yet, this proof-of-concept evaluation was limited to regularisation-based methods only and hence further experimentation is needed to fully understand the benefits of using \ac{CL} as an effective bias mitigation strategy for facial expression and action unit recognition tasks. With harder problems, as in the case of multi-label \ac{AU} detection, we see that even though regularisation-based methods achieve high accuracy, they do so by sacrificing fairness across different domain attributes. While a simplistic and naive rehearsal mechanism is able to improve model performance, our future work will aim to investigate other, more complex, pseudo-rehearsal methods~\cite{shin2017continual,churamani2020clifer,Titsias2020Functional,kim-ECCV-2020} or neuro-inspired~\cite{Kemker2018FearNetBM,Kamra2017DeepGD,Parisi2018a} on their bias mitigation abilities. 

 \begin{acronym}
\acro{AC}{Affective Computing}
\acro{ACC}{Average Accuracy}
\acro{AI}{Artificial Intelligence}
\acro{AU}{Action Unit}
\acro{AUC}{Area Under The Curve}
\acro{BP4D}{Binghamton-Pittsburgh 3D Dynamic (4D) Spontaneous Facial Expression Database}
\acro{BWT}{Backward Transfer}
\acro{CF}{Catastrophic Forgetting}
\acro{CHL}{Competitive Hebbian Learning}
\acro{CL}{Continual Learning}
\acro{Class-IL}{Class-Incremental Learning}
\acro{CLS}{Complementary Learning Systems}
\acro{CNN}{Convolutional Neural Network}
\acro{DA}{Disentangled Feature Learning}
\acro{DDC}{Domain Discriminative Classification}
\acro{DIC}{Domain Independent Classification}
\acro{Domain-IL}{Domain-Incremental Learning}
\acro{EWC}{Elastic Weight Consolidation}
\acro{FACS}{Facial Action Coding System}
\acro{FER}{Facial Expression Recognition}
\acro{FWT}{Forward Transfer}
\acro{GAN}{Generative Adversarial Network}
\acro{HRI}{Human-Robot Interaction}
\acro{MAS}{Memory Aware Synapses}
\acro{ML}{Machine Learning}
\acro{MLP}{Multi-layered Perceptron}
\acro{NC}{New Concepts}
\acro{NI}{New Instances}
\acro{NIC}{New Instances and Concepts}
\acro{NR}{Naive Rehearsal}
\acro{PRS}{Partitioning Reservoir Sampling}
\acro{RaaS}{Robotics as a Service}
\acro{RAF-DB}{Real-world Affective Faces Database}
\acro{RL}{Reinforcement Learning}
\acro{ROC}{Receiver Operating Characteristics}
\acro{SAR}{Socially Assistive Robotics}
\acro{SI}{Synaptic Intelligence}
\acro{SS}{Strategic Sampling}
\acro{Task-IL}{Task-Incremental Learning}
\end{acronym}
\balance
\bibliographystyle{IEEEtran}
\bibliography{main}
\end{document}